\documentclass[table]{ai2style/ai2}

\usepackage{microtype}
\usepackage{hyperref}
\usepackage{url}
\usepackage{booktabs, tabularx} 
\usepackage{array} 
\usepackage{graphicx}
\usepackage{lineno}
\usepackage{enumitem}
\usepackage{listings} 
\usepackage{svg}

\definecolor{ darkblue}{rgb}{0, 0, 0.5}
\hypersetup{colorlinks=true, citecolor=darkblue, linkcolor=darkblue, urlcolor=darkblue}

\usepackage{amssymb}
\usepackage{multirow}
\usepackage{bigdelim}
\usepackage{todonotes}
\usepackage{longtable}
\usepackage{tabularray}
\usepackage{wrapfig}
\usepackage[most]{tcolorbox}
\usepackage{xspace}
\usepackage[absolute]{textpos} 

\usepackage{fdsymbol}   

\usepackage[utf8]{inputenc} 
\usepackage[T1]{fontenc}    
\usepackage{amsfonts}       
\usepackage{nicefrac}       
\usepackage{amsmath}
\usepackage{csquotes}

\usepackage{siunitx}
\usepackage{arydshln}
\usepackage{soul} 

\usepackage{adjustbox}
\usepackage{pifont}
\usepackage{caption}
\usepackage{makecell}
\usepackage{subcaption}
\usepackage{bold-extra}
\usepackage{xcolor,colortbl}

\usepackage{pgf-pie}

\usepackage{placeins} 
\usepackage{etoolbox}
\preto\section{\FloatBarrier}

\usepackage{dblfloatfix}

\makeatletter
\def\fps@figure{!t}
\def\fps@table{!t}
\makeatother

\usepackage{pifont}
\usepackage{listings}

\setlist[itemize]{leftmargin=*,itemsep=0em,parsep=0.3em,topsep=0.3em}

\DeclareUnicodeCharacter{2212}{\ensuremath{-}}

\usepackage{tikz}

\usepackage{setspace}

\usepackage{nicematrix}
\newcolumntype{L}[1]{>{\raggedright\let\newline\\\arraybackslash\hspace{0pt}}m{#1}}
\newcolumntype{C}[1]{>{\centering\let\newline\\\arraybackslash\hspace{0pt}}m{#1}}
\newcolumntype{R}[1]{>{\raggedleft\let\newline\\\arraybackslash\hspace{0pt}}m{#1}}
\newcolumntype{P}[1]{>{\centering\let\newline\\\arraybackslash\columncolor{ai2lightpink}}m{#1}}
\newcommand{\huggingface}{\raisebox{-1.5pt}{\includegraphics[height=1.05em]{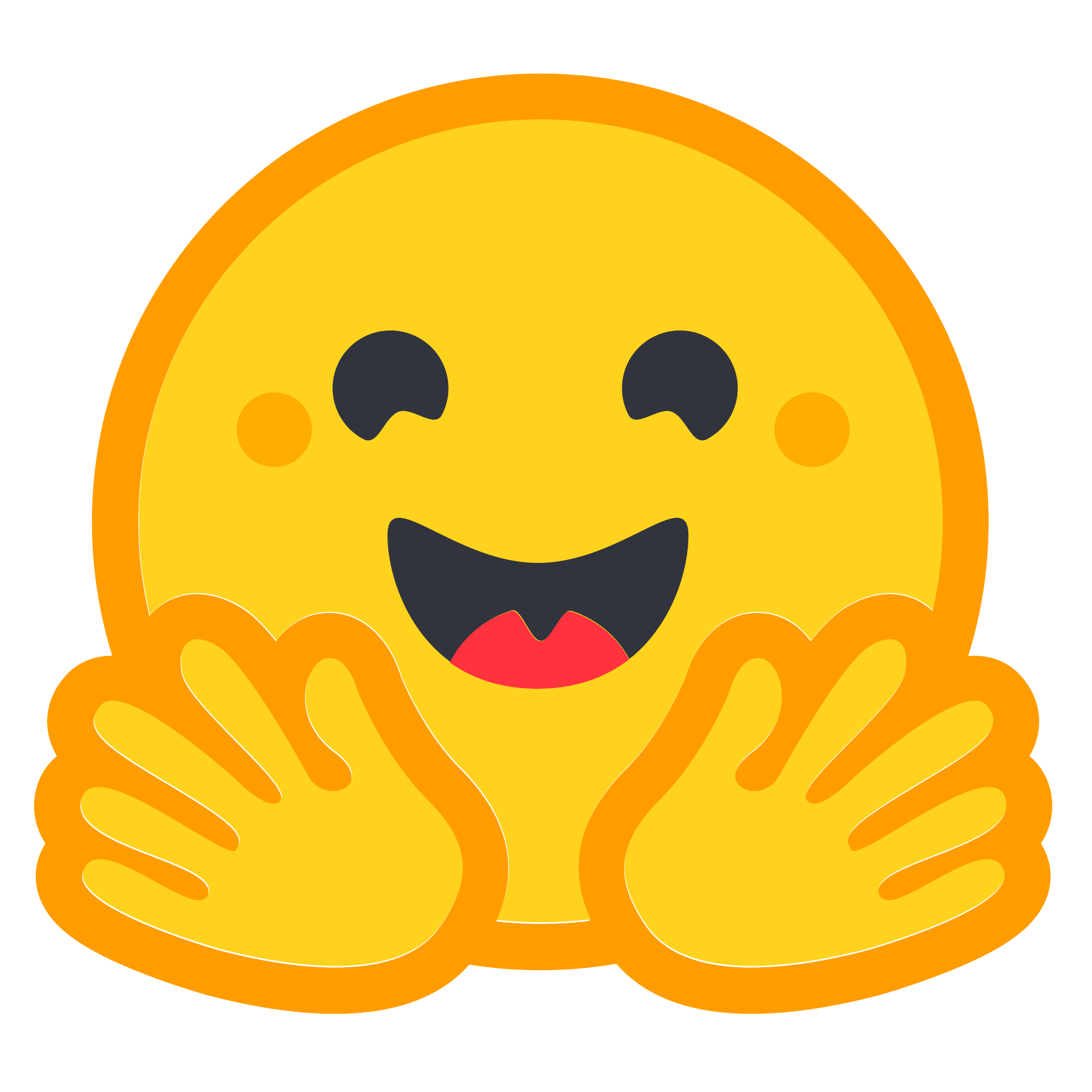}}\xspace}

\newcommand{\emailLogo}{\raisebox{-1.5pt}{\includegraphics[height=1.05em]{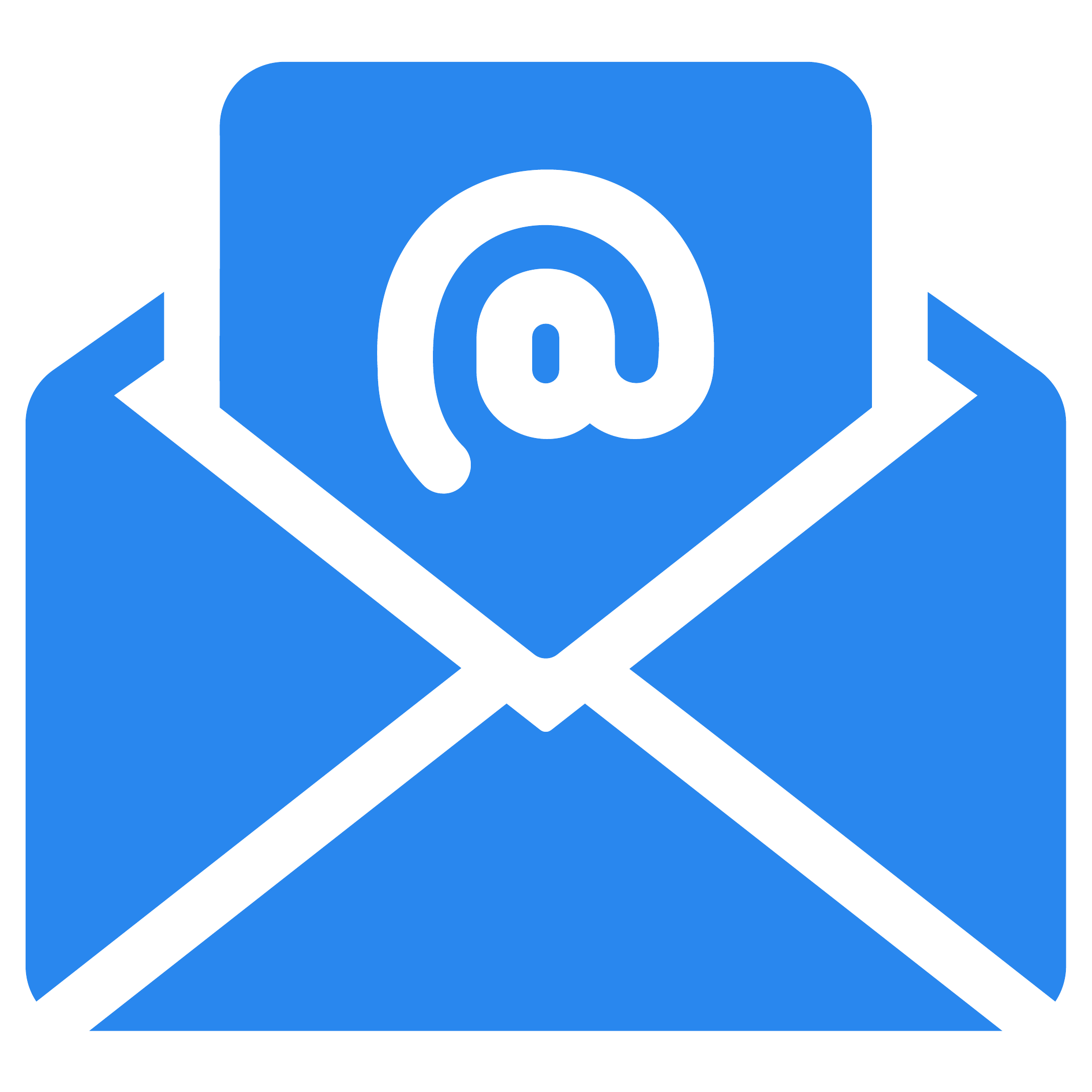}}\xspace}
\newcommand{\github}{\raisebox{-1.5pt}{\includegraphics[height=1.05em]{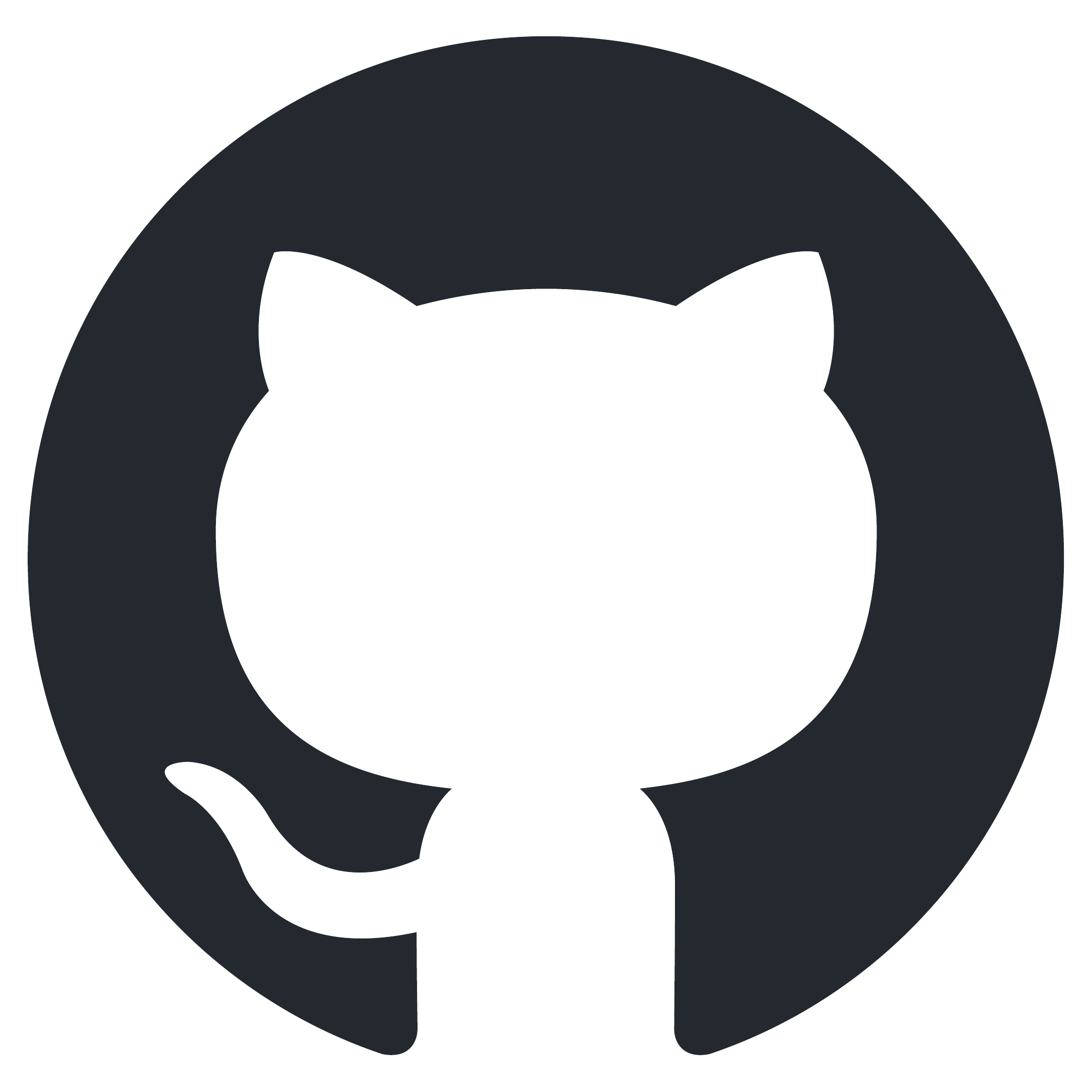}}\xspace}

\usepackage{etoolbox}
\usepackage{xspace}
\newtoggle{anon}
\togglefalse{anon}

\newcommand{\planteur}{PLANTEUR\xspace}
\newcommand{\pastis}{PASTIS\xspace}
\newcommand{\aef}{AEF\xspace}
\newcommand{\tesseravtwo}{Tessera v2\xspace}

\newcommand{\olmoearth}{OlmoEarth\xspace}
\newcommand{\olmoearthtwo}{OlmoEarth v1.2\xspace}
\newcommand{\olmoearththree}{OlmoEarth v1.3\xspace}

\def\colordefault{0a3235}
\def\colorone{f0529c}
\def\colortwo{b11be8}
\def\colorthree{6fa8dc}
\def\colorfour{0fcb8c}
\def\colorfive{12cce5}
\def\colorsix{f65834}
\def\colorgrey{B0B0B0}

\usepackage{multirow}
\usepackage{comment}
\usepackage{adjustbox}

\usepackage{tabularx}
\usepackage{graphicx}
\usepackage{tikz}
\usetikzlibrary{patterns.meta,positioning, arrows.meta,fit,calc,shapes.geometric,matrix}

\usepackage{pgfplots}
\pgfplotsset{compat=1.18}

\usepackage{pdfrender}

\usepackage{graphicx}
\makeatletter
\def\maxwidth#1{\ifdim\Gin@nat@width>#1 #1\else\Gin@nat@width\fi}
\makeatother

\usepackage{contour}
\contourlength{1pt} 
\contournumber{20}  

\usepackage{pifont}

\title{\planteur: A Crop-Type Segmentation Benchmark for Tropical Regions}

\authorTwo[]{Piper Wolters}
\authorTwo[]{Gabriel Tseng}
\authorTwo[]{Favyen Bastani}
\authorTwo[]{Joseph Redmon}
\authorThree[]{Yawen Zhang}
\authorThree[]{Hadrien Sablon}
\authorThree[]{Ando Shah}

\affiliation{Allen Institute for AI}

\abstract{%
We introduce \textbf{\planteur}, a benchmark for crop-type mapping from satellite images in the French overseas territories. Building on the \pastis design, \planteur provides georeferenced and aligned Sentinel-2, Sentinel-1, and Landsat-8 sequences with parcel-derived labels spanning six tropical regions and an 8-class label space. We define both a standard test split and a leave-one-island-out protocol to measure cross-territory generalization, and evaluate fully supervised models, linear probes, and fine-tuned foundation models. The benchmark code and data are open source and available on GitHub and Hugging Face.
}

\metadata[\quad\github Code:]{
    \href{https://github.com/allenai/planteur-benchmark}{\texttt{https://github.com/allenai/planteur-benchmark}}
}

\metadata[\vspace{.5em}\quad\huggingface Dataset:]{
    \href{https://huggingface.co/datasets/allenai/planteur-benchmark}{\texttt{https://huggingface.co/datasets/allenai/planteur-benchmark}}
}

\metadata[\vspace{.5em}\quad\emailLogo Contact:]{\color{ai2accent}{\texttt{olmoearth@allenai.org}}}

\begin{document}

\definecolor{colordefault}{HTML}{\colordefault}
\definecolor{colorgrey}{HTML}{\colorgrey}
\definecolor{colorone}{HTML}{\colorone}
\definecolor{colortwo}{HTML}{\colortwo}
\definecolor{colorthree}{HTML}{\colorthree}
\definecolor{colorfour}{HTML}{\colorfour}
\definecolor{colorfive}{HTML}{\colorfive}
\definecolor{colorsix}{HTML}{\colorsix}
\maketitle
\suppressfloats[t]
\suppressfloats[b]

\section{Introduction}

Foundation models and open data are driving rapid progress in machine learning for remote sensing \citep{garnot2021panoptic,helber2019eurosat,lacoste2023geobench,simumba2026geobench2,sumbul2019bigearthnet}. Benchmarks play a key role in tracking how model improvements carry over to real applications. Good benchmarks have: high-quality labels, associated geospatial metadata (so users can fetch additional modalities/time series), and high correlation with real-world tasks. \pastis~\citep{garnot2021panoptic} is one of the field's most trusted benchmarks because it pairs dense, high-quality parcel labels with complete geospatial metadata --- but its coverage is limited to metropolitan France. Agricultural systems where mapping could be most valuable --- smallholder, tropical, topographically complex --- are almost entirely absent from evaluation. We extend \pastis to six tropical regions to fill this gap.

We present \planteur, a new open-access crop-type segmentation benchmark. \planteur complements the existing \pastis benchmark by expanding the geographic diversity to tropical regions, while maintaining high quality labels. We use France’s Land Parcel Identification System data source for crop type labels spanning six tropical regions, and align corresponding Sentinel-2, Sentinel-1, and Landsat-8 time series.

We evaluate twelve models across linear probing, fine-tuning and supervised training, on a standard split and leave-one-island-out setup, as well as with varying inputs and hyperparameters. We find that all methods we evaluated exhibit substantially lower performance on \planteur compared to \pastis. This validates our motivation for developing \planteur and provides a new challenge for the modeling community to tackle.

To evaluate the effect of region transfer, we experiment with training on \pastis samples in addition to \planteur samples and find that those added samples, though different from the distribution of the \planteur test set, improve performance for several models. We explore performance on unseen regions by proposing a leave-one-island-out split and see a major drop in performance for most methods, suggesting a gap between a model performing well on a benchmark and generalizing to real-world tasks. 

Our contributions include 1) a dataset with permissive licenses including imagery from three sensor types, 2) geographically balanced splits, including a leave-one-island-out setup, and 3) extensive results on a variety of baseline methods, showing significant room for improvement. In Section~\ref{sec:planteur} we introduce the construction and breakdown of \planteur. In Section~\ref{sec:exp_results} we present the experimental results on \planteur across a wide range of setups.


\section{\planteur: Crop-Type Segmentation in France's Overseas Territories}
\label{sec:planteur}

\planteur is an open-access satellite image time series (SITS) semantic segmentation benchmark for crop type mapping. This dataset is publicly available at \url{https://huggingface.co/datasets/allenai/planteur-benchmark}.

\subsection{Background: \pastis}
\pastis \cite{garnot2021panoptic} comprises 2,433 multi-spectral image sequences with both semantic and instance annotations. Each sequence covers a $128\times128$ pixel patch at 10m resolution, with between 38 and 61 Sentinel-2 acquisitions taken over a one-year period, for a total of over 2 billion annotated pixels spanning more than 4,000 km$^2$ across four regions of metropolitan France. This benchmark demonstrated that parcel-based crop-type segmentation from Sentinel-2 time series can be benchmarked with high label fidelity, leveraging France's Land Parcel Identification System, which claims a relative error rate in terms of surfaces under 0.3\%~\cite{garnot2021panoptic}; this is because the parcel boundary and crop type data comes directly from farmers, and there are financial penalties and random audits that incentivize very accurate reporting. \pastis is a reliable and trustworthy benchmark, but it is geographically limited to France, a single country in Europe. Fortunately, the same high quality data exists for other French regions. \planteur extends this design to France's overseas territories, preserving the same annotation provenance while adding geographic and agronomic diversity through tropical crop regimes that are absent or underrepresented in metropolitan France, which we hope will be a useful complement to \pastis.


\subsection{Description}
\begin{figure}[t]
    \centering
    \includegraphics[width=\linewidth]{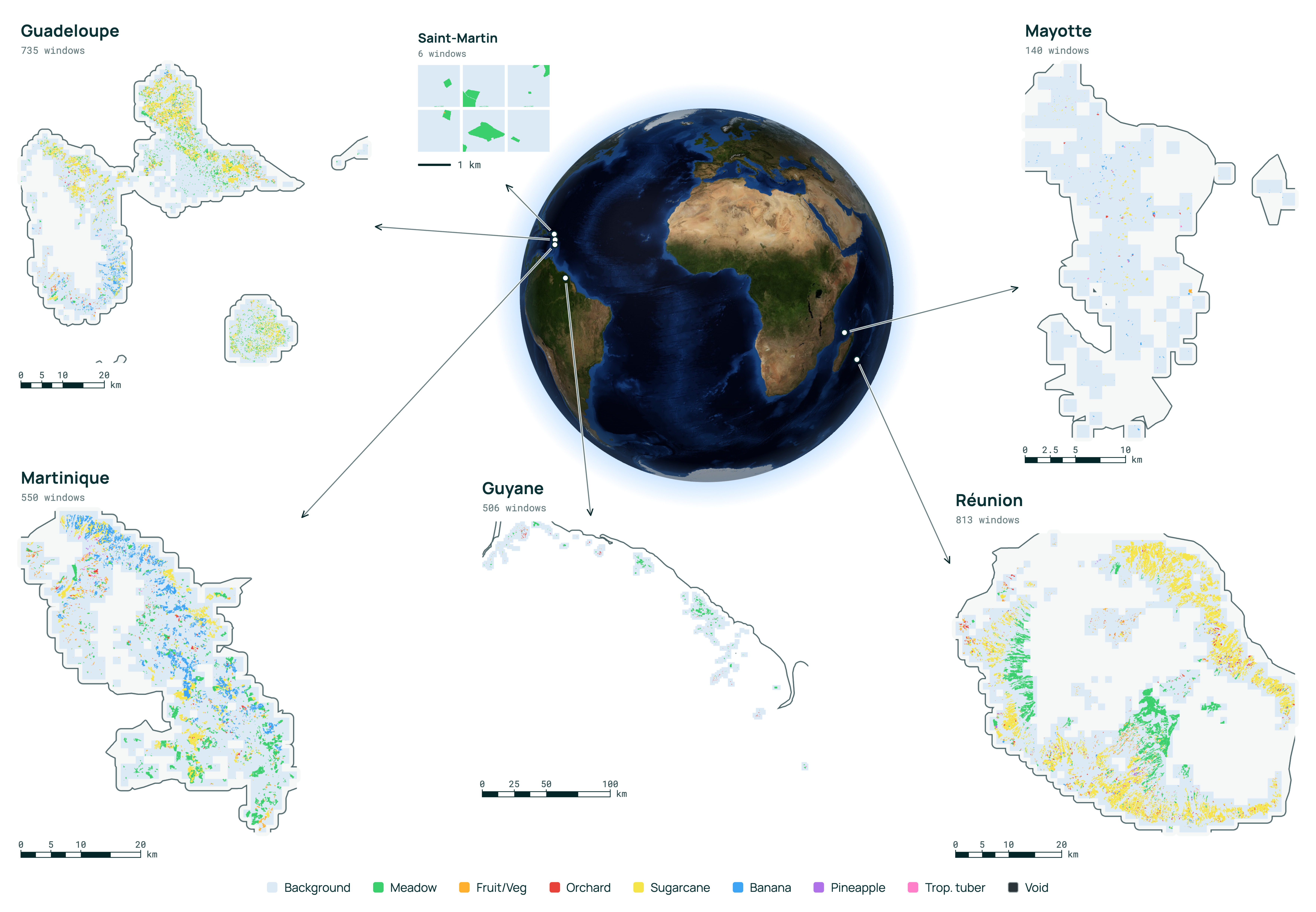}
    \caption{\planteur pixels illustrated on maps of the six tropical regions included in the dataset. Note that the whitespace represents pixels that do not include labels, whether due to being non-agricultural (e.g., mountains) or labels not existing in the data source.}
    \label{fig:planteur-map}
\end{figure}

We aim to follow the setup of \pastis for ease of adoption. \planteur consists of 2,750 multi-spectral sequences, spanning September 2018 to December 2019. Each Sentinel-2 L2A series has shape $T\times10\times128\times128$ and each Sentinel-1 RTC series $T\times2\times128\times128$, with $T$ between 40 and 150 observations (median 87 and 123 respectively; series are required to have at least 40 and are capped at 150). Landsat-8 is provided at $T\times10\times64\times64$ with a co-registered panchromatic band at $T\times1\times128\times128$, where $T$ ranges from 28 to 61 (median 30), reflecting its 16-day revisit. Like PASTIS-R \cite{pastisr}, we added additional sensor modalities to encourage multi-modal experiments, and extended the time range to make it easier to run with recent embedding products. The dataset spans 4,506 km$^2$ across R\'eunion, Guadeloupe, Martinique, Guyane, Mayotte, and Saint-Martin, geographically visualized in Figure~\ref{fig:planteur-map}.

\paragraph{Array formats.} Each patch is released as per-sensor arrays with the acquisition dates recorded in a metadata geojson, following \pastis: Sentinel-2 as $(T_{\mathrm{S2}}, 10, 128, 128)$ uint16 (B02--B08, B8A, B11, B12), Sentinel-1 as $(T_{\mathrm{S1}}, 2, 128, 128)$ float32 (VV, VH), and Landsat-8 as $(T_{\mathrm{LS}}, 10, 64, 64)$ uint16 (B1--B7, B9--B11) with the panchromatic B8 held separately at $(T_{\mathrm{LS}}, 1, 128, 128)$. Same-day acquisitions are mosaicked to a single observation, so the date lists are unique and chronological and their length matches each array's $T$.

\paragraph{Labels.} Labels are derived from the French Land Parcel Identification System for the 2019 declaration year, rasterized to the patch grid at 10\,m. Semantic labels are $(1, 128, 128)$ uint8.

Alongside the array formats provided, semantic labels are also released as georeferenced GeoTIFFs with one per patch, $128\times128$ uint8 at $10\,\mathrm{m}$ with nodata 255, keeping the patch's native UTM CRS and affine transform. The .npy arrays are pixel-identical but carry no spatial reference, so the GeoTIFFs let users rasterize additional modalities directly onto the label grid and align external data. 

\paragraph{Panoptic Segmentation Labels.}
We format the benchmark as closely to \pastis as possible, including panoptic labels. Instance annotations give a per-pixel parcel id, an object-centerness heatmap and Voronoi zones, matching the \pastis panoptic format. The focus in our experiments is on semantic segmentation and evaluating embedding models, so we leave panoptic evaluation to users interested in that task.

\subsection{Annotation}
Each pixel of \planteur is labeled as one of the $7$ crop classes or a Background, non-crop, class. The benchmark label space contains the following crop classes: Meadow, Fruit/Vegetable/Flower, Orchard, Sugarcane, Banana, Pineapple, Tropical tuber. The original label space contains additional crop types; we map any class with fewer than 40k labeled pixels to \emph{void}. Void pixels are masked out of the training loss and excluded from evaluation metrics. 

\begin{figure}[t]
    \centering
    \includegraphics[width=\linewidth]{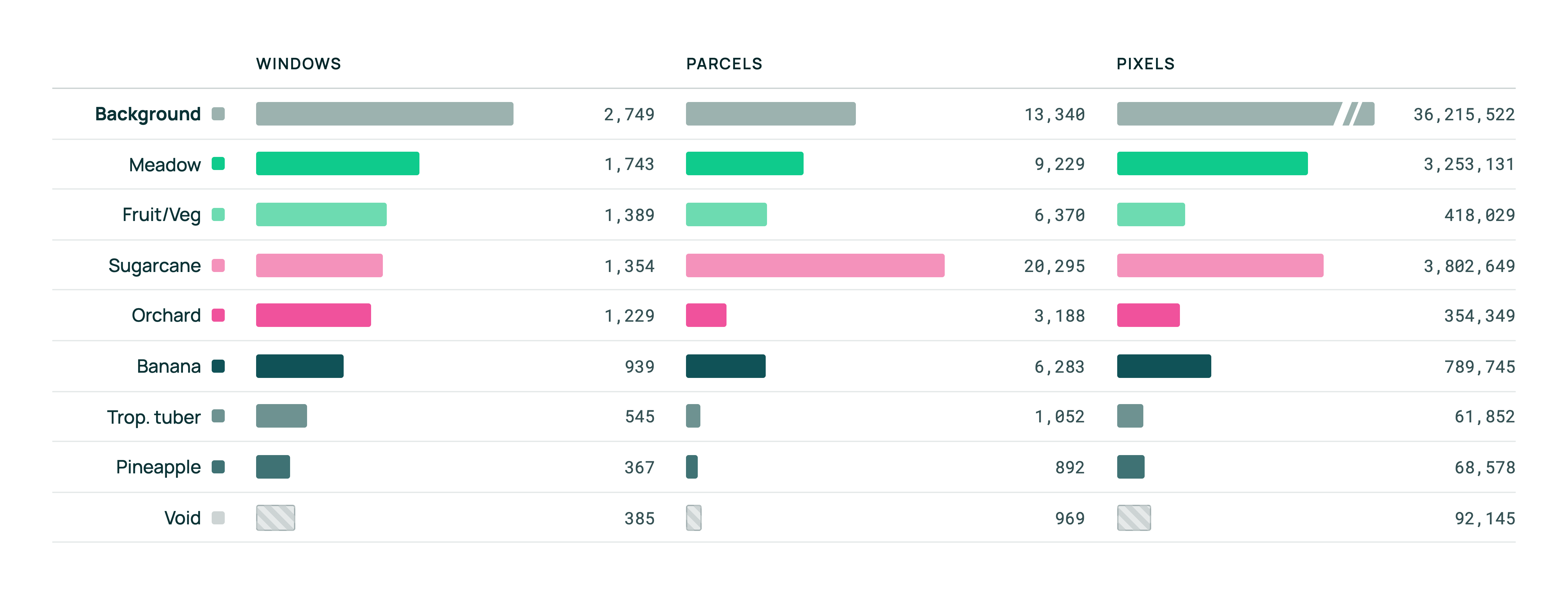}
    \caption{\planteur class distribution, in terms of number of windows containing that class, number of parcels for that class, and number of pixels labeled as that class. The background class represents non-agriculture, and the void class consists of underrepresented tropical classes.}
    \label{fig:planteur-class-distribution}
\end{figure}

\begin{figure*}[t]
      \centering
      \includegraphics[width=\linewidth]{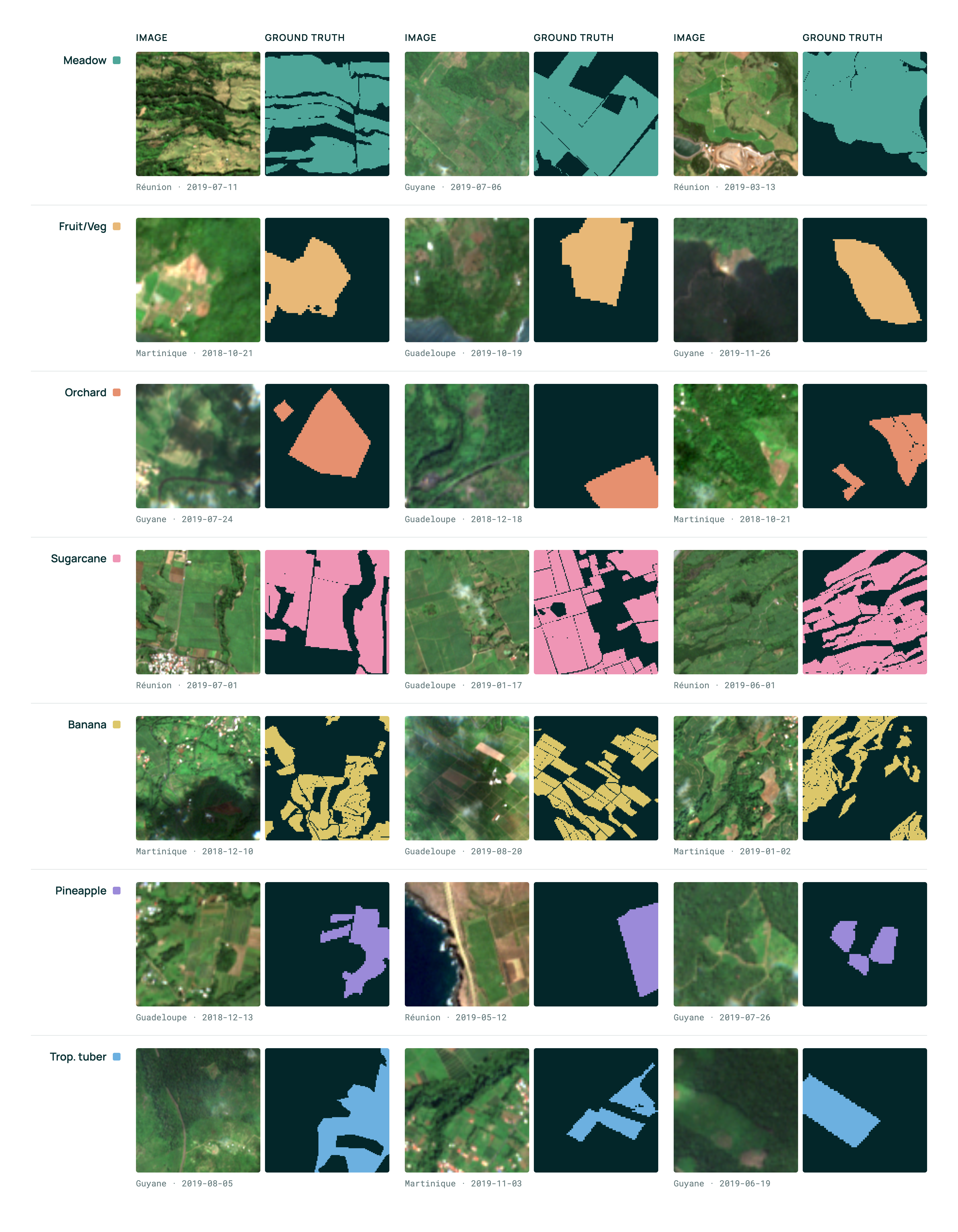}
      \caption{Example parcels for each scored crop class in \planteur. Each chip is a Sentinel-2 RGB composite from a single acquisition date, with the ground-truth mask showing \textbf{just} that class's pixels.}
      \label{fig:planteur-crop-samples}
  \end{figure*}

Class distributions, broken down by number of windows, parcels, and pixels, are shown in Figure~\ref{fig:planteur-class-distribution}. Sample images containing each crop type are in Figure~\ref{fig:planteur-crop-samples}. 

\paragraph{Void mapping.}
The following crop types are mapped to void (ignored in the loss and dropped from the metric): Vanilla, Coffee/Cacao, Corn, Potatoes, Ylang-ylang, Leguminous fodder, Grapevine, Beet, Soybeans. In total, void accounts for 92{,}145 pixels ($0.20\%$ of all pixels overall; $0.37\%$ in the test split). 

\subsection{Splits}
Splits are assigned to spatial blocks, not to individual windows. Each $128\times128$ window is mapped to the $0.1^{\circ}$ ($\approx 11$~km) latitude/longitude cell containing its centroid, and the split is assigned to the cell, so every window in a neighborhood lands in the same partition. This is necessary because the windows are densely tiled: adjacent windows observe the same landscape and crop calendar, and frequently the same parcels straddling a window boundary. Within a territory, we greedily select cells based on a hash-derived order, and assign them cumulatively until 70\%, 85\% and 100\% of that territory's windows are covered. The 70/15/15 allocation is applied independently per territory, so every territory except Saint-Martin appears in all three partitions. This yields \textbf{1{,}896 train / 411 validation / 443 test} windows (68.9\% / 14.9\% / 16.1\%); the 443-window test partition is used for all in-domain evaluation.

Because whole blocks are indivisible, realized per-territory ratios deviate from the target in inverse proportion to a territory's block count. All territories except Mayotte have \textbf{train / validation / test} ratios similar to the global ratio. Mayotte is relatively small, yielding a substantial deviation (39\% train, 15\% validation, 46\% test). Saint-Martin is \textit{very} small, so it just goes in the train set. We prioritize the spatial-block constraint over a closer ratio, because leakage would be a larger issue. Per-territory composition is given in Table~\ref{tab:planteur-perisland}.

\begin{table*}[t]
    \centering
    \small
    \setlength{\tabcolsep}{4pt}
    \begin{tabular}{l|c|c|c|c}
        Island & test windows & test pixels & Background \% & Trivial mIoU\\
        \hline
        R\'eunion & 172 & 2,805,239 & 71.02\% & 8.88\\
        Guadeloupe & 69 & 1,130,352 & 85.99\% & 10.75\\
        Martinique & 58 & 950,272 & 84.20\% & 10.52\\
        Guyane & 80 & 1,301,548 & 78.83\% & 9.85\\
        Mayotte & 64 & 1,043,953 & 99.45\% & 12.43\\
    \end{tabular}
    \caption{Per-island composition of the 443-window test split, broken down by number of windows and pixels. Note that Saint-Martin has just six windows in the train set and does not contribute to the test set.}
    \label{tab:planteur-perisland}
\end{table*}

\section{Experiments \& Results}
\label{sec:exp_results}

\subsection{Models}
We benchmark three method families on \planteur, spanning supervised models that performed well on \pastis \cite{garnot2021panoptic, garn2023vitssits}, linear probes -- released embedding products \cite{feng2025tessera, feng2026tesserav2, brown2025alphaearthfoundationsembeddingfield} and frozen encoders of remote sensing foundation models \cite{wang2025towards, astruc2024anysat, tseng2023lightweight, tseng2025galileo, olmoearthv13}, and fine-tuned foundation models \cite{tseng2026olmoearthv12, wang2025towards, astruc2024anysat, tseng2023lightweight, tseng2025galileo, feng2025tessera}.

\paragraph{Supervised from scratch.} Baselines are trained on the \planteur training split with no pretraining, using the code released for these models.
\begin{itemize}
    \item \textbf{U-TAE}~\citep{garnot2021panoptic}: U-TAE is a deep learning architecture, originally proposed alongside \pastis, that combines 2D convolutional layers with a temporal self-attention mechanism to perform highly accurate, date-agnostic spatio-temporal segmentation on satellite image time series datasets. We use the original \pastis recipe (1.08M parameters, Adam $10^{-3}$, batch 4, 100 epochs) for experiments on \planteur.
    \item \textbf{TSViT}~\citep{garn2023vitssits}: TSViT is a fully-attentional, transformer-based architecture, that was a prominent baseline model for \pastis. The model operates on $24\times24$ tiles with 12-step sequences. We use the original \pastis recipe (1.66M parameters, AdamW $10^{-3}$ with a cosine schedule and 10 warm-up epochs, batch 16, 100 epochs) for experiments on \planteur. 
\end{itemize}

Both of these models struggled with class imbalance, showing poor performance on underrepresented classes, so we experiment with different loss functions in Appendix~\ref{sec:ablations}. Experiments in the main paper use the original recipes and losses.

\paragraph{Linear probe.} We evaluate two linear-probe setups.
\begin{itemize}
    \item \textbf{Released embeddings.} For released annual embedding products, we fit a linear head on the precomputed embeddings: 50 epochs, one embedding per pixel ($\text{patch size}=1$) over a $16\times16$ window. The probe learning rate is selected from an 8-point sweep ($10^{-4}$ to $0.5$) on the validation set. At most, the learning rate affects mIoU by 2.3 points.
    \begin{itemize}
        \item \textbf{\aef}~\citep{brown2025alphaearthfoundationsembeddingfield}. AlphaEarth Foundations (AEF) is Google DeepMind's pixel-wise embedding model that distills multi-sensor satellite observations into 64-dimensional annual embeddings at 10m resolution. We use their supplied embeddings, read from precomputed rasters.
        \item \textbf{Tessera v1}\cite{feng2025tessera}. Tessera v1 is a dual-encoder model over Sentinel-2 and Sentinel-1. We use the released precomputed 128-dimensional embeddings.
        \item \textbf{Tessera v1.1}\cite{feng2025tessera}. Tessera v1.1 embeddings improve upon v1 with fewer tiling artifacts, better temporal stability, and a wider encoder. We use the released precomputed 128-dimensional embeddings.
    \end{itemize}

    \item \textbf{Frozen encoders.} Across all models we fit a linear head on features from a frozen encoder, which is never updated: 50 epochs, mean pooling, a $16\times16$ window, and a per-pixel readout obtained by bilinearly upsampling each encoder's token grid to the label resolution. The probe learning rate is selected per model from the same 8-point sweep ($10^{-4}$ to $0.5$). Patch size follows each model's native setting and is stated per model below.
    \begin{itemize}
        \item \textbf{AnySat}\cite{astruc2024anysat}. AnySat is a multi-modal JEPA-style encoder that takes in various combinations of sensors and spatial resolutions through a scale-aware patch embedding. We use a patch size of 4, window size of 16, and input up to 96 Sentinel-2 observations in 2019.
        \item \textbf{Galileo}\cite{tseng2025galileo}. Galileo is a multi-modal masked-autoencoder trained with paired global and local objectives over stacked modalities. We use a patch size of 4, window size of 16, and input 12 Sentinel-2 monthly mosaics from 2019.
        \item \textbf{Presto}\cite{tseng2023lightweight}. Presto is a lightweight transformer pretrained on pixel time series, encoding each pixel's temporal sequence independently. We use a patch size of 1, window size of 16, and input 12 Sentinel-2 monthly mosaics from 2019.
        \item \textbf{\tesseravtwo}~\citep{feng2026tesserav2}. Tessera v2 is a pixel-wise Earth foundation model producing 128-dimensional annual embeddings for 2019 at 10m. Since there are no released precomputed embeddings, we run the released v2 inference code to generate embeddings using all Sentinel-2 and Sentinel-1 observations in 2019, and keep the encoder frozen.
        \item \textbf{Copernicus-FM}\cite{wang2025towards}. Copernicus-FM is a ViT that is sensor-agnostic across spectral inputs. We resize tiles to $224\times224$ with bilinear interpolation to match the native pretraining setup; with patch size 16 and window size 128 this yields a $14\times14$ token grid per tile. It is a single-image encoder with no temporal axis so we encode each of the 12 monthly mosaics for 2019 independently and mean-pool the results.
        \item \textbf{\olmoearththree}\cite{olmoearthv13}. \olmoearththree is the \olmoearth model specifically built for embeddings. It is a $d{=}768$ encoder with a distilled linear student, and we read it out three ways --- through the distilled 64-dimensional and 128-dimensional projections, and at full register width with no projection --- reported as separate rows throughout. We use a patch size of 1, window size of 16, and 12 monthly mosaics for 2019 as input.
    \end{itemize}
\end{itemize}

\paragraph{Fine-tuned.}
\begin{itemize}
    \item \textbf{AnySat, Galileo, Presto, Tessera v1, Copernicus-FM}\cite{astruc2024anysat, tseng2025galileo, tseng2023lightweight, feng2025tessera, wang2025towards}. We also evaluate the performance of fine-tuning these foundation models. We use a window size of 128 and patch size of 16 for Copernicus-FM, a window size of 64 and patch size of 4 for \olmoearthtwo, Galileo, and AnySat and a window size of 16 and patch size of 1 for the pixelwise models, Presto and Tessera v1. The same inputs are used for fine-tuning as described for linear probes. 

    \item \textbf{\olmoearthtwo}~\citep{tseng2026olmoearthv12}. \olmoearthtwo is an update of the original \olmoearth model \cite{olmoearth}. We fine-tune the model with patch size 4 and window size 64 (unless otherwise stated) for 50 epochs, with the backbone frozen for the first 10 epochs, at $10^{-4}$ with batch size 8. We fine-tune \olmoearthtwo deliberately: v1.2 is the better fine-tuning target, whereas we released v1.3 as an embeddings-specific model.
\end{itemize}

\subsection{Protocols}
Each protocol isolates a different axis of difficulty.

\begin{itemize}
  \item \textbf{In-domain (443-window test split).} Train and test on splits of all six territories within \planteur, except Saint-Martin which is only in the train split. Measures achievable accuracy when labels exist for the target region.
  
  \item \textbf{Balanced In-domain.} Train on a subset of the in-domain train split, where the number of Background samples equals the total number of crop samples, to reduce the class imbalance challenge. Test on the in-domain test split. This is a less practical setting than the in-domain split, but allows comparing the effectiveness of different satellite image encoders.
  
  \item \textbf{Leave-one-island-out (LOIO).} Train on four territories, test on the fifth. Measures transfer to a region absent from training.

  \item \textbf{Few-shot.} Train on $X$ labeled pixels per \planteur class, $X\in\{10,25,100,1000\}$, and evaluate on the \planteur test split. Measures performance given a small number of in-domain labels.
  
  \item \textbf{Few-shot transfer.} Train on some number of \pastis samples plus $X$ labeled pixels per \planteur class, $X\in\{10,25,100,1000\}$, and evaluate on the \planteur test split. We experiment with using 1) \emph{all \pastis}, which uses all \pastis samples and adds $X$ \planteur pixels per class, and 2) \emph{balanced}, which draws $X$ pixels of every class from both \pastis and \planteur, including overlapping classes. Measures how far a small number of local labels closes the gap left by cross-region transfer.

\end{itemize}

All evaluation uses mIoU over the eight scored classes (mIoU-8); void is excluded from both the loss and the metric. For folds where a class is absent from the ground truth, we include it in the mean with IoU set to 0 to keep results comparable across territories. 
Overall pixel accuracy is reported alongside, but this metric can be deceiving due to the Background class being the majority class. 

\subsection{In-domain results (443-window test split)}
\label{sec:indomain}
Table~\ref{tab:planteur-443} reports results on the \planteur 443-window test split. Fine-tuning \olmoearthtwo is strongest at $40.82$ mIoU, ahead of both supervised baselines (TSViT $36.33$, U-TAE $35.82$). Among frozen encoders of comparable embedding dimensionality, \olmoearththree (128d) is best, followed by Tessera v2 (36.88 vs. 34.20, 2.7 points).

\begin{table*}[t]
      \centering
      \small
      \setlength{\tabcolsep}{4pt}
      \begin{tabular}{l|l|l|S[table-format=2.2]|S[table-format=2.2]}
      Family & \textit{} & Model & \multicolumn{1}{c|}{mIoU} & \multicolumn{1}{c}{acc.}\\
      \hline
      \multirow{11}{*}{\shortstack[l]{Linear\\probe}} & \multirow{3}{*}{\textit{Released embeddings}} & \aef & 27.68 & 89.50\\
       &  & Tessera v1.1 & 32.78 & {\bfseries 89.84}\\
       &  & Tessera v1 & {\bfseries 33.13} & 89.77\\
      \cline{2-5}
       & \multirow{8}{*}{\textit{Frozen encoders}} & AnySat & 28.79 & 86.59\\
       &  & Galileo & 34.71 & 89.91\\
       &  & Presto & 16.25 & 80.15\\
       &  & Copernicus-FM & 26.46 & 86.74\\
       &  & \tesseravtwo & 34.20 & 90.26\\
       &  & OlmoEarth v1.3 (64d) & 35.17 & {\bfseries 90.55}\\
       &  & OlmoEarth v1.3 (128d) & 36.88 & 90.46\\
       &  & OlmoEarth v1.3 (768d) & {\bfseries 38.94} & 90.12\\
      \hline
          \multirow{8}{*}{Supervised} &  & U-TAE & 35.82 & 90.62\\
           &  & TSViT & 36.33 & 90.27\\
           &  & AnySat & 38.35 & 90.58\\
           &  & Galileo & 38.25 & 91.27\\
           &  & Presto & 23.69 & 87.36\\
           &  & Tessera v1 & 35.77 & 89.97\\
           &  & Copernicus-FM & 32.54 & 89.21\\
           &  & OlmoEarth v1.2 FT & \cellcolor{colorone!25}{\bfseries 40.82} & \cellcolor{colorone!25}{\bfseries 91.80}\\
          \hline
          \multicolumn{3}{l|}{Trivial (all Background)} & 10.07 & 80.56\\
      \end{tabular}
      \caption{Results on the \planteur 443-window test split. Best within each block in bold, best overall highlighted.}
      \label{tab:planteur-443}
\end{table*}

\begin{figure}[t]
    \centering
    \includegraphics[width=\linewidth]{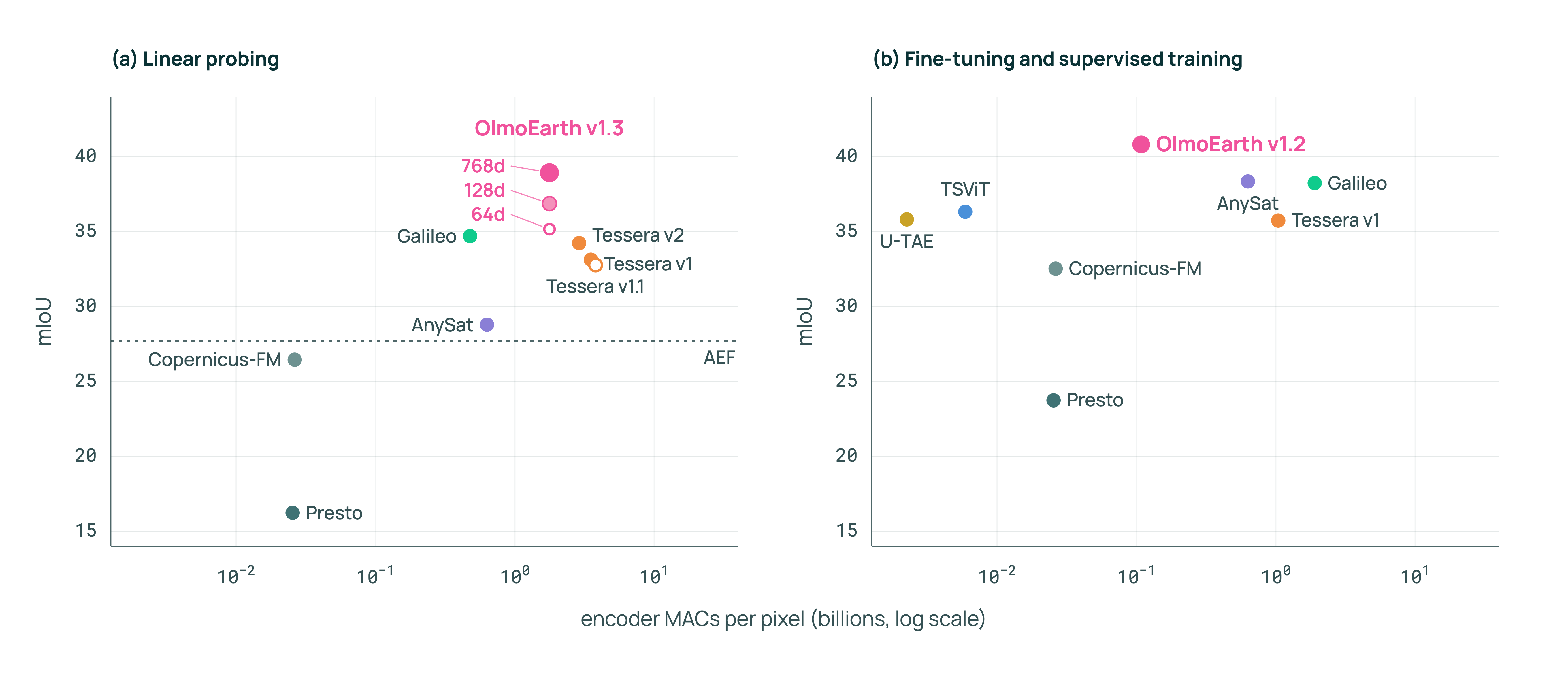}
    \caption{MACs vs. mIoU on \planteur standard test set, separated into two plots for linear probe and supervised methods. The reported MACs represent compute cost of forward pass, so released embedding products wouldn't bear this cost. The dotted line marks \aef, a closed-weight embedding product for which encoder MACs are unavailable.}
    \label{fig:macs-vs-mious}
\end{figure}

\paragraph{Monthly mosaics vs. all observations.} We evaluate both models that input monthly mosaics (e.g. Galileo, OlmoEarth) and models that input all observations (e.g. AnySat, Tessera). These models cannot be compared apples-to-apples due to compute cost, so we plot performance as a function of the compute cost in Figure~\ref{fig:macs-vs-mious}. This shows that \olmoearthtwo is the best model at a reasonable cost, compared to AnySat and Tessera v1 which have lower performance and a higher compute cost.

\paragraph{Do additional input modalities improve performance?} For the \olmoearth variants, we experiment with varying the input modalities. Results for both \olmoearththree probes and \olmoearthtwo fine-tuning with Sentinel-2, Sentinel-1, and Landsat-8 input combinations are reported in Table~\ref{tab:planteur-443-olmoearth-inputs}. Sentinel-1 combined with Sentinel-2 gives a slight increase (768d: +0.5, 128d: +0.2, \olmoearthtwo: +0.8 mIoU points), while adding Landsat-8 does not consistently help, and no combination improves the 64-dimensional embeddings.

\begin{table*}[t]
      \centering
      \small
      \setlength{\tabcolsep}{4pt}
      \begin{tabular}{l|c|c|c|c|c}
          Model & S2 & S1 & S1+S2 & S2+LS & S1+S2+LS\\
          \hline
            OlmoEarth v1.3 (64d) & \textbf{35.17} & 24.64 & 34.95 & 34.90 & 35.01\\
            OlmoEarth v1.3 (128d) & 36.88 & 24.45 & \textbf{37.04} & 36.68 & 36.02\\
            OlmoEarth v1.3 (768d) & 38.94 & 26.91 & \textbf{39.39} & 38.50 & 39.03\\
            \hline
            OlmoEarth v1.2 FT & 40.82 & 32.41 & \textbf{41.62} & 40.42 & 41.01\\
      \end{tabular}
      \caption{\olmoearth varying inputs on the 443-window \planteur test split. Best input combination per row in bold.}
      \label{tab:planteur-443-olmoearth-inputs}
\end{table*}

\begin{figure}[t]
    \centering
    \includegraphics[width=\linewidth]{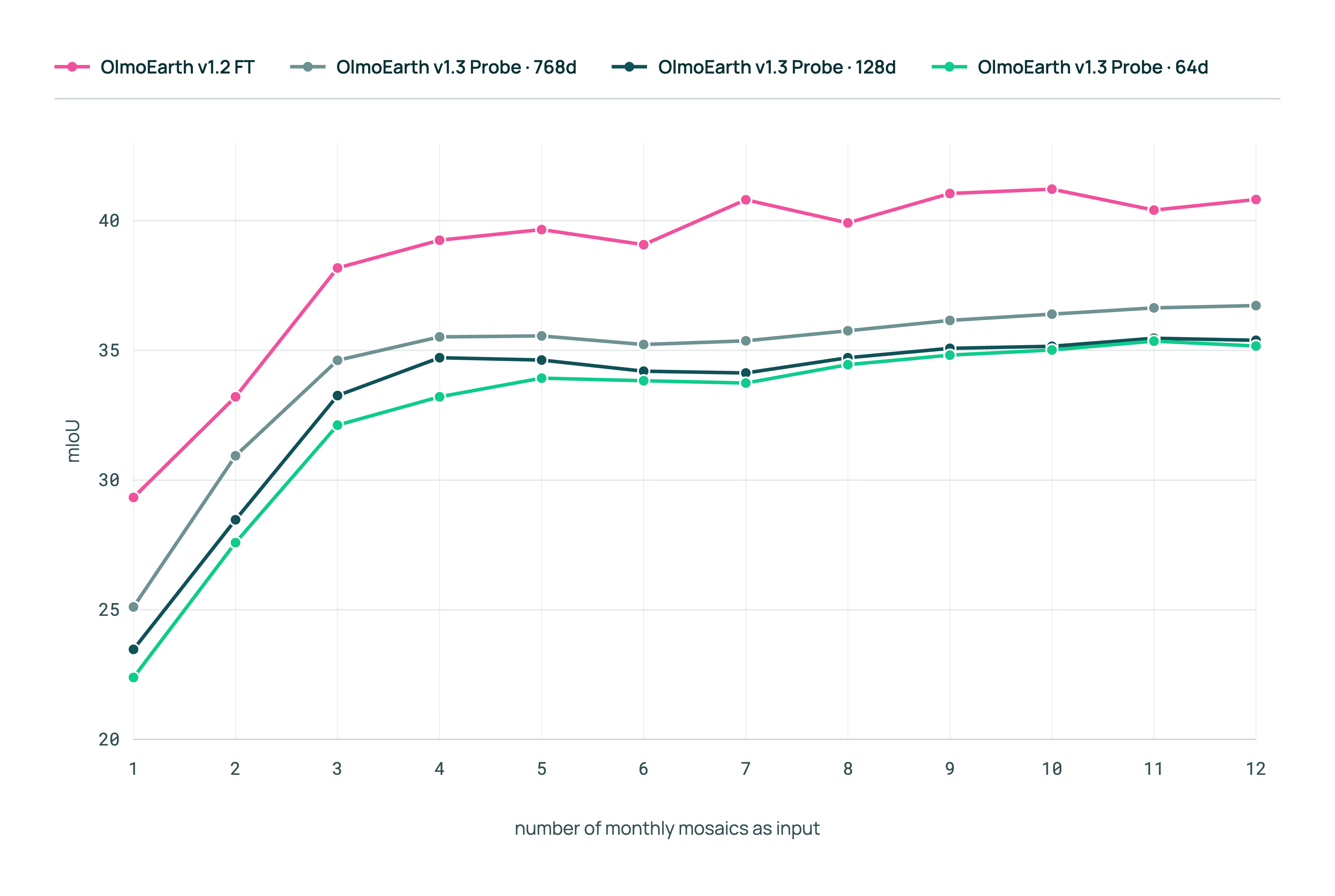}
    \caption{Temporal sequence-length ablation for \olmoearththree linear probes and \olmoearthtwo fine-tuning: mIoU-8 as a function of the number of monthly mosaics provided as input. Performance improves with a larger sequence length, though the rate of increase is largest between a sequence length of one and three for all methods. \olmoearththree probes in this ablation use a fixed learning rate of 0.1 rather than a validation-selected sweep.}
    \label{fig:temporal-sequence-ablation}
\end{figure}

\paragraph{How important is temporal context?} Results for varying the temporal sequence length for the \olmoearththree linear probes and \olmoearthtwo fine-tuning are shown in Figure~\ref{fig:temporal-sequence-ablation}. Across all three \olmoearththree embedding dimension sizes, performance increases by 11.6–12.8 mIoU points from one to twelve monthly mosaics, with the largest jump in improvement between a sequence length of one and three.

\paragraph{Does window size matter?} We vary the window size for the frozen encoder linear probe results with all three dimensionality variants of \olmoearththree. Table~\ref{tab:planteur-window-size-ablation} shows that a larger window size of 32 increases mIoU by 1.3–1.7 points over the $w=16$ setting of Table~\ref{tab:planteur-443}, for both the distilled read-outs and the full width. Fine-tuning \olmoearthtwo shows the same trend (35.82 at $w=4$ to 40.82 at $w=64$).

\begin{table*}[t]
      \centering
      \small
      \setlength{\tabcolsep}{4pt}
      \begin{tabular}{l|c|c|c|c|c}
            Model & $w=4$ & $w=8$ & $w=16$ & $w=32$ & $w=64$\\
            \hline
              OlmoEarth v1.3 (64d) & 33.21 & 34.51 & 35.17 & 36.82 & \textbf{37.56}\\
              OlmoEarth v1.3 (128d) & 34.69 & 37.59 & 36.88 & \textbf{38.20} & 37.10\\
              OlmoEarth v1.3 (768d) & 35.92 & 36.16 & 38.94 & \textbf{40.20} & 39.36\\
              \hline
              OlmoEarth v1.2 FT & 35.82 & 38.67 & 39.81 & 40.65 & \textbf{40.82}\\
      \end{tabular}
      \caption{Spatial-context ablation on the 443-window \planteur test split. Best window size per row in bold.}
      \label{tab:planteur-window-size-ablation}
\end{table*}

\subsection{Balanced In-domain}
\label{sec:balanced}
The train / validation / test split used in Section~\ref{sec:indomain} has majority Background samples, and this class imbalance is likely the reason for models over-predicting Background. This also means that employing specialized training strategies for imbalanced datasets can provide substantial performance gains --- however, this is not the focus of our benchmark. Thus, we create a class-balanced training set to better distinguish the effectiveness of different remote sensing foundation models. We use the same test set to keep the evaluation realistic.

Confusion matrices for models trained on both the imbalanced and balanced sets are shown in Figure~\ref{fig:planteur-confusion-matrix-pairs} in the appendix. These matrices and the results in Table~\ref{tab:planteur-downweighted} show an improvement (but not a \textit{huge} improvement) for all methods, except for U-TAE, when using the balanced train set instead of the baseline train split. There are still a lot of pixels predicted as Background rather than the ground truth crop, suggesting that the benchmark's difficulty goes beyond class imbalance.

\begin{table*}[t]
    \centering
    \small
    \setlength{\tabcolsep}{4pt}
    \begin{tabular}{l|c|c}
        Model & baseline & balanced\\
        \hline
          OlmoEarth v1.3 (64d) & 35.17 & 36.74\\
          OlmoEarth v1.3 (128d) & 36.88 & 39.13\\
          OlmoEarth v1.3 (768d) & \textbf{38.94} & \textbf{40.91}\\
        \tesseravtwo & 34.20 & 37.42\\
        \aef & 27.68 & 28.71\\
        \hline
          U-TAE & 35.82 & 33.98\\
          TSViT & 36.33 & 37.95\\
          OlmoEarth v1.2 FT & \cellcolor{colorone!25}\textbf{40.82} & \cellcolor{colorone!25}\textbf{41.94}\\
    \end{tabular}
    \caption{Baseline vs. balanced, where the balanced setup reduces the number of Background samples to equal the total number of crop samples, evaluated on \planteur 443-window test split. Best within each block in bold, best overall highlighted. The difference in performance between columns is not large, suggesting that the task difficulty is beyond dealing with class imbalance.}
    \label{tab:planteur-downweighted}
\end{table*}

\subsection{Leave-one-island-out (LOIO)}
\label{sec:loio}
Held-out-territory results are in Table~\ref{tab:planteur-loio}. The best mean across folds is $25.28$ (\olmoearthtwo fine-tuned), followed by the \olmoearththree 128-dimensional probe (24.13, 1.2 mIoU points lower).

\begin{table*}[t]
    \centering
    \small
    \setlength{\tabcolsep}{4pt}

    \begin{tabular}{l|c|c|c|c|c|c}
        Model & R\'eunion & Guadeloupe & Martinique & Guyane & Mayotte & mean\\
        \hline
        OlmoEarth v1.3 (64d) & 22.70 & 30.38 & 29.93 & 14.12 & \textbf{12.49} & 21.92\\
        OlmoEarth v1.3 (128d) & \textbf{26.43} & 31.16 & \textbf{35.16} & \textbf{15.50} & 12.42 & \textbf{24.13}\\
        OlmoEarth v1.3 (768d) & 13.84 & \cellcolor{colorone!25}\textbf{33.90} & 32.69 & 13.95 & 12.48 & 21.37\\
        \tesseravtwo & 11.50 & 28.79 & 29.57 & 12.26 & 12.43 & 18.91\\
        \aef & 9.88 & 23.08 & 20.77 & 12.75 & 12.42 & 15.78\\
        \hline
        U-TAE & 22.34 & 28.43 & 30.25 & 14.52 & \cellcolor{colorone!25}\textbf{12.52} & 21.61\\
        TSViT & 18.72 & 28.33 & 33.86 & 12.23 & 12.37 & 21.10\\
        OlmoEarth v1.2 FT & \cellcolor{colorone!25}\textbf{27.77} & \textbf{32.87} & \cellcolor{colorone!25}\textbf{36.54} & \cellcolor{colorone!25}\textbf{16.82} & 12.42 & \cellcolor{colorone!25}\textbf{25.28}\\
        \hline
        Trivial (all Background) & 9.33 & 10.17 & 10.18 & 10.33 & 12.42 & 10.49\\
    \end{tabular}

    \caption{\planteur leave-one-island-out (LOIO) results. Mean is across the 5 folds. Best within each block in bold, best overall highlighted.}
    \label{tab:planteur-loio}
\end{table*}

\paragraph{LOIO compared to in-domain.} This setup is difficult for all models. Every method achieves roughly 55--65\% of its in-domain performance: likely due to the crop type distribution varying widely across the different islands and some islands only having a subset of the crop classes in \planteur.

\paragraph{Interpreting Mayotte.} Mayotte is majority Background, so results on this fold do not hold much weight. We do not throw this region out though, because a small region with very few positive samples is a realistic application. None of the reported methods have a meaningful performance on this island.

\subsection{Few-shot: \planteur labels only}
\label{sec:fewshot}
We test few-shot learning on \planteur by training on $X$ pixels per class, $X\in\{10,25,100,1000\}$, and evaluating on the \planteur test set, see Table~\ref{tab:planteur-fewshot}. For the supervised models, we use random cropping around the labeled pixels to try to minimize overfitting.

\begin{table*}[t]
    \centering
    \small
    \setlength{\tabcolsep}{4pt}
    \begin{tabular}{l|c|c|c|c}
        Model & $X=10$ & $X=25$ & $X=100$ & $X=1000$\\
        \hline
          OlmoEarth v1.3 (64d) & 20.57 & 21.06 & 25.23 & 27.67\\
          OlmoEarth v1.3 (128d) & \textbf{21.80} & \textbf{21.50} & 26.96 & 28.88\\
          OlmoEarth v1.3 (768d) & 20.75 & 20.96 & \textbf{27.57} & \cellcolor{colorone!25}\textbf{29.66}\\
          \tesseravtwo & 18.61 & 18.28 & 23.51 & 27.23\\
          \aef & 16.16 & 17.63 & 21.93 & 24.25\\
          \hline
          U-TAE & 5.84 & 8.14 & 14.49 & 22.59\\
          TSViT & 9.71 & 11.32 & 18.32 & 27.44\\
          OlmoEarth v1.2 FT & \cellcolor{colorone!25}\textbf{22.05} & \cellcolor{colorone!25}\textbf{22.03} & \cellcolor{colorone!25}\textbf{28.83} & \textbf{28.35}\\
        \hline
        Trivial (all Background) & \multicolumn{4}{c}{10.07}\\
    \end{tabular}
    \caption{Few-shot on \planteur labels only: train on $X$ labeled \planteur pixels per class, evaluate on the 443-window test split. Best within each block in bold, best overall highlighted.}
    \label{tab:planteur-fewshot}
\end{table*}

\paragraph{Performance with \textit{very} few examples.} At $X=10$, the models have just eighty labeled pixels, and there is a wide range of performance. \olmoearththree (128d) outperforms Tessera v2 and AEF by 3.2 and 5.6 mIoU points, and the best supervised model (TSViT) by 12.1 points. With just 10 or 25 examples, the supervised models overfit and perform similarly to predicting Background everywhere. There are likely better training recipes that would improve performance for supervised models.

\paragraph{Comparing OlmoEarth v1.3 embeddings (linear probe) and OlmoEarth v1.2 (fine-tuning).} In the few-shot setting, \olmoearthtwo fine-tuning and \olmoearththree linear probes are roughly tied at small $X$ (22.05 vs.\ 21.80 at $X=10$, 22.03 vs.\ 21.50 at $X=25$). \olmoearthtwo leads by 1.3 points at $X=100$, and the 768d probe leads by 1.3 at $X=1000$. Linear probing on frozen embeddings matches full fine-tuning when labels are scarce,  
  at a fraction of the cost.


\subsection{Few-shot transfer: adding \pastis}
\label{sec:fewshot-transfer}
\begin{figure*}[t]
    \centering
    \includegraphics[width=\linewidth]{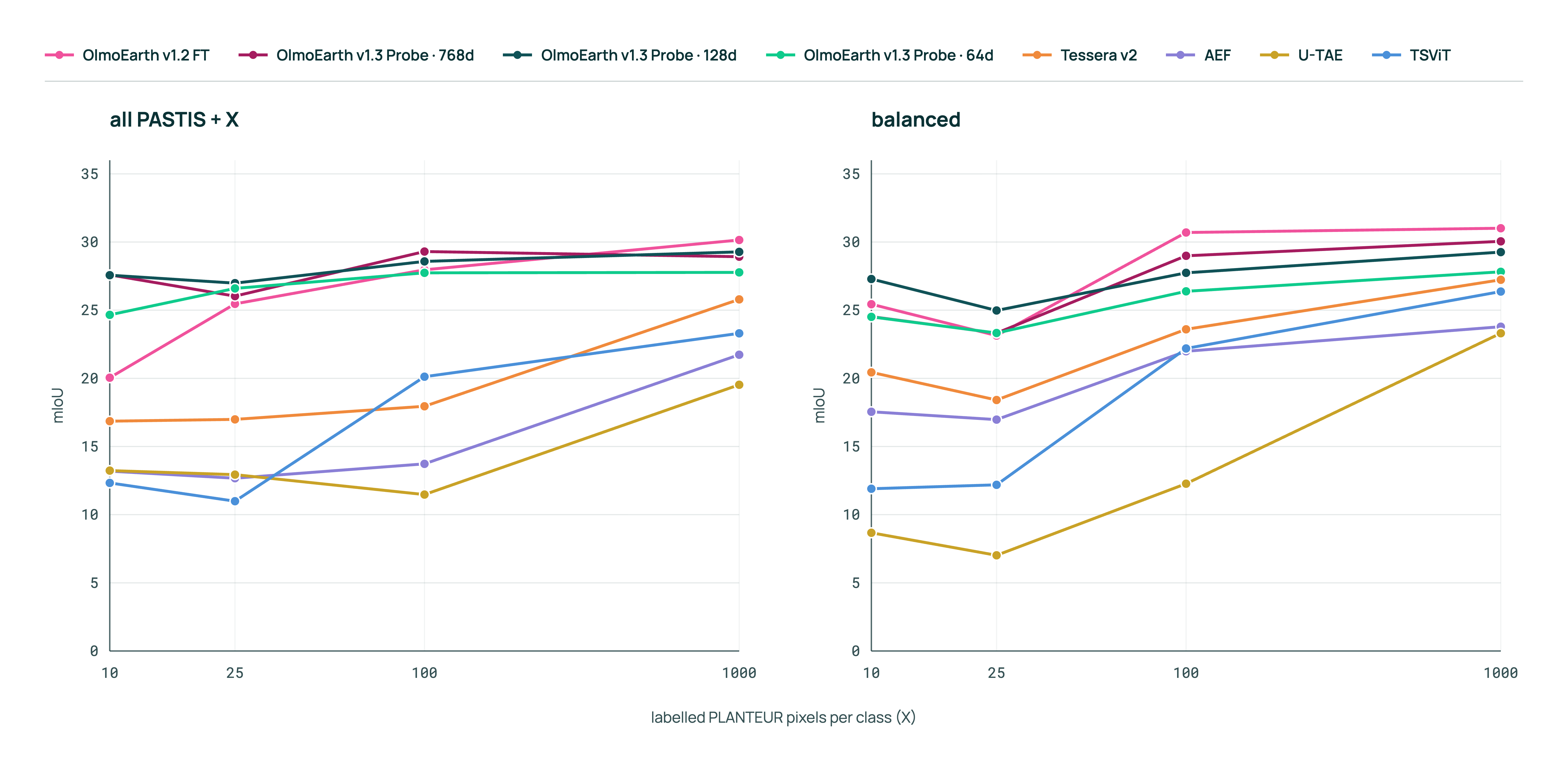}
    \caption{\planteur few-shot transfer: mIoU-8 against $X$ labeled \planteur pixels per class, on top of \pastis.}
    \label{fig:planteur-xshot-results}
\end{figure*}

Next, we see if adding \pastis data to the training distribution, using the same \planteur budget as Section~\ref{sec:fewshot}, can narrow the remaining gap between using the whole training split and the few-shot setting. Like Section~\ref{sec:fewshot}, we apply random cropping to the supervised runs. Results for all models are shown in Figure~\ref{fig:planteur-xshot-results}, which visualizes the results when using $X$ \planteur pixels per class, \emph{all \pastis}, which uses all \pastis samples and adds $X$ \planteur pixels per class, and \emph{balanced}, which draws $X$ pixels of every class from both \pastis and \planteur, including overlapping classes.

\paragraph{All \pastis versus balanced.} When the full \pastis training set is added to the $X$ labeled \planteur pixels per class, the \planteur supervision is a tiny fraction of the training signal: at most $8X$ pixels, against millions of labeled \pastis pixels. The four tropical classes (sugarcane, banana, pineapple, and tropical tuber) appear only in \planteur, so most models never learn to predict them at small $X$, and their mIoU comes almost entirely from the classes the two datasets share. Their scores then reflect how well \pastis-trained features happen to transfer to those shared classes rather than how many \planteur labels were added, which explains the sometimes non-monotonic trends in the \emph{all \pastis} rows.

When there is a more equal distribution of crops between \pastis and \planteur, the tropical classes carry real weight and so performance generally rises from $X=100$ to $X=1000$, though most methods dip between $X=10$ and $X=25$.

\paragraph{Do \pastis samples improve performance on \planteur when limited samples are available?} 
Compared to the few-shot results in Table~\ref{tab:planteur-fewshot}, all \olmoearththree variants see an improvement when adding \pastis samples in both the \emph{all \pastis} and \emph{balanced} settings, with a consistent boost for most values of $X$. Tessera v2 and AEF do not see the same increase; the \emph{balanced} mix barely changes them, and \emph{all \pastis} lowers them (−5.6 and −8.1 at $X=100$).

For \olmoearthtwo fine-tuning, the \emph{balanced} mix helps at every $X$ (+1.1 to +3.4 mIoU), while \emph{all \pastis} is mixed (−2.0 to +3.4), consistent with the tropical classes being swamped by \pastis supervision. 

\section{Conclusion}
\label{sec:conclusion}
We introduce PLANTEUR, a new crop-type mapping segmentation benchmark, meant to expand the community's evaluations to tropical regions, while maintaining the high quality and accuracy of the ground truth labels. As the results show, this is a difficult benchmark compared to PASTIS. Results across models are not saturated, so there is room to improve. We hope the community tackles this benchmark with new ideas and methods! 

Although PLANTEUR expands data coverage to tropical crops and regions, while maintaining high accuracy in parcel boundaries and crop types, a limitation of France's Land Parcel Identification System is completeness -- participation of farmers in reporting their data to this system is only mandatory if they participate in the EU Common Agricultural Policy, so there are still many small-scale farms that don't participate in this program and thus do not report to this system. This is evident in the original data analysis, where we had to throw out crops like coffee and vanilla because the number of samples was far too low. Also, this program combines some tropical fields into generic or vague categories (e.g., mango and lychee fields are lumped into the orchards category). There is also a chance of crop change in between seasons that is not accounted for, or differences in growing seasons (e.g., R\'eunion is in the southern hemisphere so the growing seasons are out of sync with the exported imagery). In future work, it is critical to source additional high quality crop labels that expand the distribution to more and more small-scale crops.

\clearpage
\bibliographystyle{ieeenat_fullname}
\bibliography{main}

@String(CVPR= {IEEE Conf. Comput. Vis. Pattern Recog.})

@String(CVPR  = {CVPR})

@misc{brown2025alphaearthfoundationsembeddingfield,
      title={{AlphaEarth Foundations}: An embedding field model for accurate and efficient global mapping from sparse label data}, 
      author={Christopher F. Brown and Michal R. Kazmierski and Valerie J. Pasquarella and William J. Rucklidge and Masha Samsikova and Chenhui Zhang and Evan Shelhamer and Estefania Lahera and Olivia Wiles and Simon Ilyushchenko and Noel Gorelick and Lihui Lydia Zhang and Sophia Alj and Emily Schechter and Sean Askay and Oliver Guinan and Rebecca Moore and Alexis Boukouvalas and Pushmeet Kohli},
      year={2025},
      eprint={2507.22291},
      archivePrefix={arXiv},
      primaryClass={cs.CV},
      url={https://arxiv.org/abs/2507.22291}, 
}

@article{feng2025tessera,
  title={TESSERA: Temporal Embeddings of Surface Spectra for Earth Representation and Analysis},
  author={Feng, Zhengpeng and Jaffer, Sadiq and Knezevic, Jovana and Sormunen, Silja and Young, Robin and Lisaius, Madeline and Immitzer, Markus and Ball, James and Atzberger, Clement and Coomes, David A and others},
  journal={arXiv preprint arXiv:2506.20380},
  year={2025}
}

@misc{feng2026tesserav2,
      title={TESSERA v2: Scaling Pixel-wise Earth Foundation Models}, 
      author={Z. Feng and S. Jaffer and I. Shokar and J. Knezevic and M. Elvers and C. Atzberger and R. Young and A. Naik and N. Robinson and A. Blake and D. Coomes and A. Madhavapeddy and S. Keshav},
      year={2026},
      eprint={2607.03949},
      archivePrefix={arXiv},
      primaryClass={cs.CV},
      url={https://arxiv.org/abs/2607.03949}, 
}

@inproceedings{garn2023vitssits,
  title={ViTs for SITS: Vision Transformers for Satellite Image Time Series},
  author={Garn, Michael and others},
  booktitle={Proceedings of the IEEE/CVF Conference on Computer Vision and Pattern Recognition (CVPR)},
  year={2023}
}

@article{tseng2023lightweight,
  title={Lightweight, pre-trained transformers for remote sensing timeseries},
  author={Tseng, Gabriel and Cartuyvels, Ruben and Zvonkov, Ivan and Purohit, Mirali and Rolnick, David and Kerner, Hannah},
  journal={arXiv preprint arXiv:2304.14065},
  year={2023}
}

@article{helber2019eurosat,
  title={Eurosat: A novel dataset and deep learning benchmark for land use and land cover classification},
  author={Helber, Patrick and Bischke, Benjamin and Dengel, Andreas and Borth, Damian},
  journal={IEEE Journal of Selected Topics in Applied Earth Observations and Remote Sensing},
  volume={12},
  number={7},
  pages={2217--2226},
  year={2019},
  publisher={IEEE}
}

@inproceedings{garnot2021panoptic,
  title={Panoptic segmentation of satellite image time series with convolutional temporal attention networks},
  author={Garnot, Vivien Sainte Fare and Landrieu, Loic},
  booktitle={Proceedings of the IEEE/CVF International Conference on Computer Vision},
  pages={4872--4881},
  year={2021}
}

@article{astruc2024anysat,
  title={{AnySat: An Earth} Observation Model for Any Resolutions, Scales, and Modalities},
  author={Astruc, Guillaume and Gonthier, Nicolas and Mallet, Clement and Landrieu, Loic},
  journal={arXiv preprint arXiv:2412.14123},
  year={2024}
}

@inproceedings{sumbul2019bigearthnet,
  title={Bigearthnet: A large-scale benchmark archive for remote sensing image understanding},
  author={Sumbul, Gencer and Charfuelan, Marcela and Demir, Beg{\"u}m and Markl, Volker},
  booktitle={IGARSS 2019-2019 IEEE International Geoscience and Remote Sensing Symposium},
  pages={5901--5904},
  year={2019},
  organization={IEEE}
}

@inproceedings{
tseng2025galileo,
title={Galileo: Learning Global \& Local Features of Many Remote Sensing Modalities},
author={Gabriel Tseng and Anthony Fuller and Marlena Reil and Henry Herzog and Patrick Beukema and Favyen Bastani and James R Green and Evan Shelhamer and Hannah Kerner and David Rolnick},
booktitle={Forty-second International Conference on Machine Learning},
year={2025},
}

@inproceedings{wang2025towards,
  title={Towards a unified copernicus foundation model for earth vision},
  author={Wang, Yi and Xiong, Zhitong and Liu, Chenying and Stewart, Adam J and Dujardin, Thomas and Bountos, Nikolaos Ioannis and Zavras, Angelos and Gerken, Franziska and Papoutsis, Ioannis and Leal-Taix{\'e}, Laura and others},
  booktitle={Proceedings of the IEEE/CVF International Conference on Computer Vision},
  year={2025}
}

@inproceedings{olmoearth,
  title={OlmoEarth: Stable Latent Image Modeling for Multimodal Earth Observation},
  author={Herzog, Henry and Bastani, Favyen and Zhang, Yawen and Tseng, Gabriel and Redmon, Joseph and Sablon, Hadrien and Park, Ryan and Morrison, Jacob and Buraczynski, Alexandra and Farley, Karen and others},
  booktitle={CVPR},
  year={2026}
}

@article{tseng2026olmoearthv12,
  title={OlmoEarth v1.2: A more efficient family of OlmoEarth models},
  author={Tseng, Gabriel and Zhang, Yawen and Bastani, Favyen and Herzog, Henry and Redmon, Joseph and Sablon, Hadrien and Wolters, Piper and Shah, Ando and Johnson, Patrick Alan and Wilhelm, Christopher and Beukema, Patrick},
  journal={arXiv preprint arXiv:2605.20804},
  year={2026}
}

@article{olmoearthv13,
  title={OlmoEarth v1.3: Embeddings},
  author={Tseng, Gabriel and Zhang, Yawen and Bastani, Favyen and Redmon, Joseph and Sablon, Hadrien and Wolters, Piper and Shah, Ando and Johnson, Patrick},
  journal={},
  year={2026}
}

@article{pastisr,
  author       = {Vivien Sainte Fare Garnot and
                  Lo{\"{\i}}c Landrieu and
                  Nesrine Chehata},
  title        = {Multi-Modal Temporal Attention Models for Crop Mapping from Satellite
                  Time Series},
  journal      = {CoRR},
  volume       = {abs/2112.07558},
  year         = {2021},
  url          = {https://arxiv.org/abs/2112.07558},
  eprinttype   = {arXiv},
  eprint       = {2112.07558},
  bibsource    = {dblp computer science bibliography, https://dblp.org}
}

@inproceedings{lacoste2023geobench,
  title     = {GEO-Bench: Toward Foundation Models for Earth Monitoring},
  author    = {Lacoste, Alexandre and Lehmann, Nils and Rodriguez, Pau and Crastici, Evan and Segler, Benjamin and Matcham, Hannah and Courville, Aaron and David, David and Longepe, Nicolas and Gava, Alexandre and others},
  booktitle = {Advances in Neural Information Processing Systems 36 (NeurIPS Datasets and Benchmarks Track)},
  volume    = {36},
  pages     = {51080--51093},
  year      = {2023},
  publisher = {Curran Associates, Inc.}
}

@article{simumba2026geobench2,
  title={GEO-Bench-2: From Performance to Capability, Rethinking Evaluation in Geospatial AI},
  author={Simumba, Naomi and Lehmann, Nils and Fraccaro, Paolo and Alemohammad, Hamed and De Mel, Geeth and Khan, Salman and Maskey, Manil and Long{\'e}p{\'e}, Nicolas and Zhu, Xiao Xiang and Kerner, Hannah and others},
  journal={Transactions on Machine Learning Research},
  year={2026}
}

\clearpage
\section{Appendix}
\label{sec:appendix}

\subsection{Ablations}
\label{sec:ablations}

\subsubsection{Class weighting and focal loss}
Table~\ref{tab:planteur-loss-ablations} compares class weighting and focal loss against the default losses of Table~\ref{tab:planteur-443}. Inverse-frequency weighting hurts every model, by 4.0--8.8 mIoU points. The other variants change the supervised baselines by at most 1.9 points: U-TAE is best with inverse-square-root weighting (35.82 $\rightarrow$ 36.85) and TSViT with focal loss alone (36.33 $\rightarrow$ 36.45). \olmoearthtwo is best with inverse-square-root weighting (40.82 $\rightarrow$ 41.10) and loses 0.6--7.7 points under every other alternative. The exception is the \olmoearththree (768d) linear probe, which gains 2.5 points (38.94 $\rightarrow$ 41.43) with inverse-square-root weighting. Because no single loss helps consistently, we keep the default losses for all models.

\begin{table*}[t]
      \centering
      \small
      \setlength{\tabcolsep}{4pt}
      \begin{tabular}{l|l|c|c|c|c|c}
          Model & Regime & baseline & inv & invsqrt & focal $\gamma=2$ & invsqrt+focal\\
          \hline
          U-TAE & supervised & 35.82 & 28.82 & \textbf{36.85} & 36.51 & 36.68\\
          TSViT & supervised & 36.33 & 27.55 & 35.76 & \textbf{36.45} & 34.48\\
          \hline
          \olmoearthtwo & fine-tune & 40.82 & 33.16 & \textbf{41.10} & 40.21 & 39.34\\
          \olmoearththree (768d) & linear probe & 38.94 & 34.93 & \textbf{41.43} & 38.61 & 40.49\\
      \end{tabular}
      \caption{Class-weighting and focal-loss ablations on \planteur (443-window test split), mIoU-8. Best per row in bold. }
      \label{tab:planteur-loss-ablations}
\end{table*}

\begin{table*}[t]
    \centering
    \small
    \setlength{\tabcolsep}{4pt}
    \begin{tabular}{l|l|c|c|c|c}
        Model & Arm & $X=10$ & $X=25$ & $X=100$ & $X=1000$\\
        \hline
        OlmoEarth v1.3 (64d) & all \pastis & 24.64 & 26.61 & 27.73 & 27.77\\
        OlmoEarth v1.3 (64d) & balanced & 24.51 & 23.31 & 26.39 & 27.78\\
        OlmoEarth v1.3 (128d) & all \pastis & \cellcolor{colorone!25}\textbf{27.57} & \cellcolor{colorone!25}\textbf{26.98} & 28.58 & 29.29\\
        OlmoEarth v1.3 (128d) & balanced & 27.26 & 24.98 & 27.74 & 29.18\\
        OlmoEarth v1.3 (768d) & all \pastis & 27.54 & 26.04 & \textbf{29.30} & 28.92\\
        OlmoEarth v1.3 (768d) & balanced & 24.55 & 23.31 & 29.00 & \textbf{30.04}\\
        \tesseravtwo & all \pastis & 16.65 & 17.01 & 17.96 & 25.79\\
        \tesseravtwo & balanced & 20.44 & 18.41 & 23.60 & 27.23\\
        \aef & all \pastis & 13.39 & 12.95 & 13.79 & 21.73\\
        \aef & balanced & 17.55 & 17.36 & 22.21 & 23.88\\
        \hline
        U-TAE & all \pastis & 13.23 & 12.94 & 11.47 & 19.53\\
        U-TAE & balanced & 8.67 & 7.02 & 12.27 & 23.31\\
        TSViT & all \pastis & 12.33 & 10.99 & 20.12 & 23.30\\
        TSViT & balanced & 11.90 & 12.19 & 22.19 & 26.37\\
        OlmoEarth v1.2 FT & all \pastis & 20.05 & \textbf{25.46} & 27.96 & 30.15\\
        OlmoEarth v1.2 FT & balanced & \textbf{25.44} & 23.14 & \cellcolor{colorone!25}\textbf{30.70} & \cellcolor{colorone!25}\textbf{31.01}\\
        \hline
        Trivial & -- & \multicolumn{4}{c}{10.07}\\
    \end{tabular}
    \caption{Unseen-region transfer: \pastis plus $X$ labeled \planteur pixels per class. \emph{all \pastis} keeps the full \pastis training set; \emph{balanced} draws $X$ pixels of every class from both sources. Best within each block in bold, best overall highlighted.}
    \label{tab:planteur-transfer-mix}
\end{table*}

\subsubsection{Few-Shot Plus Unseen Region Transfer}
Table~\ref{tab:planteur-transfer-mix} shows the results for the few-shot \planteur plus unseen region transfer experiments where we add \pastis samples to the training set before evaluating on the \planteur test set. These results correspond to Figure~\ref{fig:planteur-xshot-results} in the main paper.

\subsubsection{Unseen Region Transfer}
As the proposed benchmark is a regional extension of an existing one, it is interesting to ask whether there can be transfer between the two. We fine-tune \olmoearththree and \olmoearthtwo on \pastis, then fit a linear probe on \planteur and evaluate on the \planteur test split. There are four overlapping classes between \pastis and \planteur and thus four unseen (aside from training the linear probe) classes introduced in \planteur. The results, shown in Table~\ref{tab:unseen-transfer}, are not impressive, which is why we emphasize the positive results of few-shot transfer, as discussed in Section~\ref{sec:fewshot-transfer}, where adding just a few examples of your region of interest to a much larger data source, can result in strong performance.

\begin{table*}[t]
      \centering
      \small
      \setlength{\tabcolsep}{4pt}
      \begin{tabular}{l|c|c|c}
          Model & mIoU-8 & trop-4 & acc.\\
          \hline
      OlmoEarth v1.3 (64d) & 28.28 & 22.53 & 89.05\\
      OlmoEarth v1.3 (128d) & 31.93 & 30.86 & 87.53\\
      OlmoEarth v1.3 (768d) & 35.82 & 33.99 & 90.73\\
      \hline
      OlmoEarth v1.2 FT & 37.83 & 35.65 & 90.13\\
      \end{tabular}
      \caption{Unseen transfer: encoders fine-tuned on \pastis, then linear-probed on the \planteur 443-window test split. The table shows mIoU-8 and mIoU over just the four tropical classes absent from \pastis (Sugarcane, Banana, Pineapple, Tropical tuber; trop-4). Each cell is the best of an 8-point learning-rate sweep selected on validation.}
      \label{tab:unseen-transfer}
\end{table*}

\subsubsection{LOIO Ablations}
We repeat a few of the ablations mentioned in Section~\ref{sec:indomain} for the LOIO protocol.

\paragraph{Do additional input modalities improve performance in this setup?} We again run experiments with varying input modalities for \olmoearththree and \olmoearthtwo. Input modality does not have a consistent effect on LOIO results as shown in Table~\ref{tab:planteur-loio-olmoearth-inputs}.

\begin{table*}[t]
    \centering
    \small
    \setlength{\tabcolsep}{4pt}
    \begin{tabular}{l|l|c|c|c|c|c|c}
          Model & Inputs & R\'eunion & Guadeloupe & Martinique & Guyane & Mayotte & mean\\
          \hline                                                OlmoEarth v1.3 (64d) & S2 & 22.70 & \textbf{30.38} & \textbf{29.93} & \textbf{14.12} & 12.49 & \textbf{21.92}\\
            OlmoEarth v1.3 (64d) & S1 & 15.08 & 21.39 & 22.31 & 11.86 & 12.41 & 16.61\\
            OlmoEarth v1.3 (64d) & S1+S2 & 20.72 & 30.10 & 29.45 & 14.00 & 12.43 & 21.34\\
            OlmoEarth v1.3 (64d) & S2+LS & \textbf{22.98} & 30.03 & 28.92 & 13.46 & 12.61 & 21.60\\
            OlmoEarth v1.3 (64d) & S1+S2+LS & 21.27 & 29.99 & 29.10 & 14.03 & 12.41 & 21.36\\
            \hline
            OlmoEarth v1.3 (128d) & S2 & 26.43 & \textbf{31.16} & 35.16 & 15.50 & 12.42 & 24.13\\
            OlmoEarth v1.3 (128d) & S1 & 14.49 & 21.99 & 23.18 & 11.79 & 12.40 & 16.77\\
            OlmoEarth v1.3 (128d) & S1+S2 & \cellcolor{colorone!25}\textbf{27.12} & 30.99 & \textbf{35.27} & \textbf{15.73} & 12.42 & \textbf{24.31}\\
            OlmoEarth v1.3 (128d) & S2+LS & 25.71 & 30.66 & 34.15 & 14.78 & 12.42 & 23.54\\
            OlmoEarth v1.3 (128d) & S1+S2+LS & 18.76 & 30.70 & 34.41 & 15.40 & 12.42 & 22.34\\
            \hline
            OlmoEarth v1.3 (768d) & S2 & 13.84 & \cellcolor{colorone!25}\textbf{33.90} & 32.69 & 13.95 & 12.48 & 21.37\\
            OlmoEarth v1.3 (768d) & S1 & 13.53 & 21.86 & 23.34 & 13.12 & 12.30 & 16.83\\
            OlmoEarth v1.3 (768d) & S1+S2 & \textbf{14.20} & 33.40 & \textbf{36.08} & 14.43 & 12.42 & \textbf{22.11}\\
            OlmoEarth v1.3 (768d) & S2+LS & 13.40 & 32.42 & 34.14 & 14.72 & 12.45 & 21.43\\
            OlmoEarth v1.3 (768d) & S1+S2+LS & 13.91 & 32.39 & 34.30 & \textbf{15.58} & 12.42 & 21.72\\
            \hline
            OlmoEarth v1.2 FT & S2 & 22.67 & 30.60 & \cellcolor{colorone!25}\textbf{38.25} & 15.26 & 12.61 & 23.88\\
            OlmoEarth v1.2 FT & S1 & 18.38 & 24.87 & 26.96 & 12.42 & 12.24 & 18.97\\
            OlmoEarth v1.2 FT & S1+S2 & 24.59 & \textbf{30.62} & 35.59 & 15.75 & 12.43 & 23.80\\
            OlmoEarth v1.2 FT & S2+LS & 24.11 & 29.99 & 36.43 & 15.48 & 12.42 & 23.69\\
            OlmoEarth v1.2 FT & S1+S2+LS & \textbf{25.65} & 29.29 & 37.42 & \cellcolor{colorone!25}\textbf{17.07} & 12.70 & \cellcolor{colorone!25}\textbf{24.43}\\
    \end{tabular}
    \caption{Leave-one-island-out (LOIO) input ablations for \olmoearth. Best mIoU-8 within each read-out block in bold, best overall highlighted, except for the Mayotte column which is not meaningful.}
    \label{tab:planteur-loio-olmoearth-inputs}
\end{table*}

\paragraph{What is the optimal window size for LOIO?} We run experiments to vary the spatial context for the \olmoearththree models. Results in Table~\ref{tab:planteur-loio-window-size} are mixed: a larger window size improves performance on most of the held out islands for the 64d and 128d probes, but not for the 768d probe.

\begin{table*}[t]
      \centering
      \small
      \setlength{\tabcolsep}{4pt}
        \begin{tabular}{l|l|c|c|c|c|c|c}
              Probe & $w$ & R\'eunion & Guadeloupe & Martinique & Guyane & Mayotte & mean\\
              \hline
              OlmoEarth v1.3 (64d) & 4 & 20.57 & 29.29 & 26.78 & 11.70 & 12.40 & 20.15\\
              OlmoEarth v1.3 (64d) & 8 & 20.42 & 29.12 & 29.65 & 13.48 & 12.52 & 21.04\\
              OlmoEarth v1.3 (64d) & 16 & 22.70 & 30.38 & 29.93 & 14.12 & 12.49 & \textbf{21.92}\\
              \hline
              OlmoEarth v1.3 (128d) & 4 & 20.25 & 29.48 & 28.10 & 13.43 & 12.42 & 20.74\\
              OlmoEarth v1.3 (128d) & 8 & 19.63 & 30.83 & 32.30 & 13.35 & 12.53 & 21.73\\
              OlmoEarth v1.3 (128d) & 16 & 26.43 & 31.16 & 35.16 & 15.50 & 12.42 & \textbf{24.13}\\
              \hline
              OlmoEarth v1.3 (768d) & 4 & 12.99 & 30.77 & 29.21 & 12.77 & 12.42 & 19.63\\
              OlmoEarth v1.3 (768d) & 8 & 16.60 & 33.22 & 33.96 & 14.12 & 12.57 & \textbf{22.09}\\
              OlmoEarth v1.3 (768d) & 16 & 13.84 & 33.90 & 32.69 & 13.95 & 12.48 & 21.37\\
        \end{tabular}
      \caption{Spatial-context ablation under LOIO protocol. Mean mIoU-8 is across the 5 folds. Best mean per section in bold.}
      \label{tab:planteur-loio-window-size}
\end{table*}

\subsubsection{Overall Accuracies}
In the main paper we mostly report mIoU as it is the prioritized metric. Overall accuracy can also tell us useful information, so we include Tables~\ref{tab:planteur-443-olmoearth-inputs-acc}, \ref{tab:planteur-window-size-ablation-acc}, \ref{tab:planteur-downweighted-acc}, \ref{tab:planteur-fewshot-acc}, and \ref{tab:planteur-loio-accuracy} of overall accuracies for each experimental setup in the main paper.
  \begin{table*}[t]
        \centering
        \small
        \setlength{\tabcolsep}{4pt}
        \begin{tabular}{l|c|c|c|c|c}
            Model & S2 & S1 & S1+S2 & S2+LS & S1+S2+LS\\
            \hline
              OlmoEarth v1.3 (64d) & \textbf{90.55} & 84.64 & 90.21 & 90.47 & 90.26\\
              OlmoEarth v1.3 (128d) & 90.46 & 85.76 & 90.47 & 90.30 & \textbf{90.61}\\
              OlmoEarth v1.3 (768d) & 90.12 & 86.66 & \textbf{90.34} & 89.97 & 90.20\\
              \hline
              OlmoEarth v1.2 FT & \textbf{91.80} & 89.49 & 91.63 & 91.50 & 91.40\\
        \end{tabular}
        \caption{\olmoearth varying inputs on the 443-window \planteur test split, overall pixel accuracy (\%). Same runs as Table~\ref{tab:planteur-443-olmoearth-inputs}. Best input combination per row in
  bold.}
        \label{tab:planteur-443-olmoearth-inputs-acc}
  \end{table*}

    \begin{table*}[t]
        \centering
        \small
        \setlength{\tabcolsep}{4pt}
        \begin{tabular}{l|c|c|c|c|c}
              Model & $w=4$ & $w=8$ & $w=16$ & $w=32$ & $w=64$\\
              \hline
                OlmoEarth v1.3 (64d) & 89.85 & 90.24 & \textbf{90.55} & 90.16 & 89.55\\
                OlmoEarth v1.3 (128d) & 90.28 & 90.36 & \textbf{90.46} & 90.13 & 90.06\\
                OlmoEarth v1.3 (768d) & 88.89 & 91.01 & 90.12 & \textbf{91.18} & 91.09\\
                \hline
                OlmoEarth v1.2 FT & 89.52 & 90.32 & 90.67 & 91.39 & \textbf{91.80}\\
        \end{tabular}
        \caption{Spatial-context ablation on the 443-window \planteur test split, overall pixel accuracy (\%). Same runs as Table~\ref{tab:planteur-window-size-ablation}. Best window size per row in bold.}
        \label{tab:planteur-window-size-ablation-acc}
  \end{table*}

  \begin{table*}[t]
      \centering
      \small
      \setlength{\tabcolsep}{4pt}
      \begin{tabular}{l|c|c}
          Model & baseline & balanced\\
          \hline
            OlmoEarth v1.3 (64d) & \textbf{90.55} & 89.02\\
            OlmoEarth v1.3 (128d) & 90.46 & 90.18\\
            OlmoEarth v1.3 (768d) & 90.12 & \textbf{90.20}\\
          \tesseravtwo & 90.26 & 89.69\\
          \aef & 89.50 & 87.74\\
          \hline
            U-TAE & 90.62 & 89.86\\
            TSViT & 90.27 & 88.88\\
          OlmoEarth v1.2 FT & \cellcolor{colorone!25}\textbf{91.80} & \cellcolor{colorone!25}\textbf{90.24}\\
      \end{tabular}
      \caption{Baseline vs.\ balanced, overall pixel accuracy (\%), where balanced reduces the number of Background samples to equal the total number of crop samples, evaluated on the \planteur 443-window test
  split. Same runs as Table~\ref{tab:planteur-downweighted}. Best within each block in bold, best overall highlighted.}
      \label{tab:planteur-downweighted-acc}
  \end{table*}

  \begin{table*}[t]
      \centering
      \small
      \setlength{\tabcolsep}{4pt}
      \begin{tabular}{l|c|c|c|c}
          Model & $X=10$ & $X=25$ & $X=100$ & $X=1000$\\
          \hline
          OlmoEarth v1.3 (64d) & 69.16 & \cellcolor{colorone!25}\textbf{58.42} & 68.51 & 70.59\\
          OlmoEarth v1.3 (128d) & \cellcolor{colorone!25}\textbf{69.99} & 56.91 & \cellcolor{colorone!25}\textbf{69.60} & 70.82\\
          OlmoEarth v1.3 (768d) & 57.76 & 51.71 & 66.04 & 68.33\\
          \tesseravtwo & 59.06 & 47.76 & 62.58 & 71.89\\
          \aef & 49.09 & 49.14 & 64.53 & \textbf{72.56}\\
          \hline
          U-TAE & 26.70 & 35.54 & 50.95 & 60.12\\
          TSViT & 34.72 & 32.49 & 55.77 & \cellcolor{colorone!25}\textbf{74.60}\\
          OlmoEarth v1.2 FT & \textbf{63.41} & \textbf{56.96} & \textbf{68.58} & 71.62\\
          \hline
          Trivial (all Background) & \multicolumn{4}{c}{80.56}\\
      \end{tabular}
      \caption{Few-shot on \planteur labels only, overall pixel accuracy (\%): train on $X$ labeled \planteur pixels per class with no \pastis data, evaluate on the 443-window test split. Same runs as Table~\ref{tab:planteur-fewshot}. Best within each block in bold, best overall highlighted.}
      \label{tab:planteur-fewshot-acc}
  \end{table*}

  \begin{table*}[t]
    \centering
    \small
    \setlength{\tabcolsep}{4pt}
    \begin{tabular}{l|c|c|c|c|c|c}
        Model & R\'eunion & Guadeloupe & Martinique & Guyane & Mayotte & mean\\
        \hline
        OlmoEarth v1.3 (64d) & 83.84 & 87.92 & 87.85 & 83.31 & 99.35 & 88.45\\
        OlmoEarth v1.3 (128d) & \textbf{84.98} & 87.28 & 88.56 & \textbf{84.48} & \cellcolor{colorone!25}\textbf{99.39} & \textbf{88.94}\\
        OlmoEarth v1.3 (768d) & 75.03 & \textbf{88.73} & \textbf{89.12} & 82.75 & 99.25 & 86.98\\
        \tesseravtwo & 75.18 & 87.00 & 88.23 & 82.97 & 99.26 & 86.53\\
        \aef & 74.29 & 85.89 & 85.28 & 80.20 & 99.37 & 85.01\\
        \hline
        U-TAE & 83.66 & 87.51 & 88.35 & 83.71 & 97.78 & 88.20\\
        TSViT & 80.73 & 85.82 & 88.07 & 81.84 & 98.46 & 86.98\\
        OlmoEarth v1.2 FT & \cellcolor{colorone!25}\textbf{85.76} & \cellcolor{colorone!25}\textbf{89.06} & \cellcolor{colorone!25}\textbf{90.07} & \cellcolor{colorone!25}\textbf{84.77} & \textbf{99.35} & \cellcolor{colorone!25}\textbf{89.80}\\
        \hline
        Trivial (all Background) & 74.64 & 81.36 & 81.44 & 82.64 & 99.36 & 83.89\\
    \end{tabular}
    \caption{\planteur leave-one-island-out (LOIO), Sentinel-2 inputs: pixel accuracy (\%). Mean is across the 5 folds. Best within each block in bold, best overall highlighted.}
    \label{tab:planteur-loio-accuracy}
\end{table*}

\subsection{Additional Visuals}
\begin{figure*}[t]
    \centering
    \includegraphics[width=\textwidth,keepaspectratio]{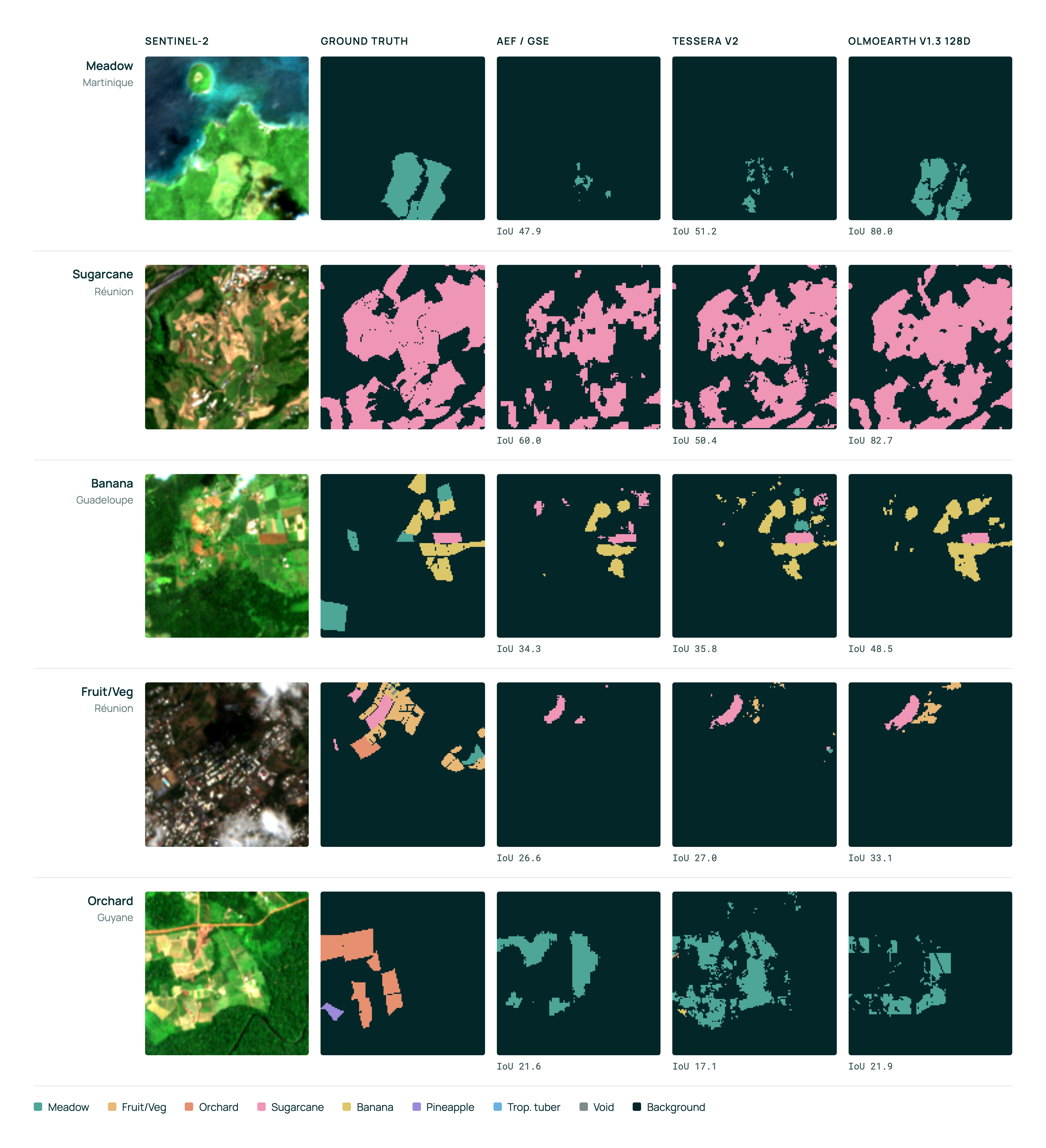}
    \caption{Qualitative \planteur examples, with each row showing a Sentinel-2 mosaic, corresponding ground truth labels, and predicted mask from linear probing AEF, Tessera v2, and \olmoearththree.}
    \label{fig:planteur-predictions}
\end{figure*}

\paragraph{Sample Predictions.} Figure~\ref{fig:planteur-predictions} shows samples of Sentinel-2 image, ground truth mask, and model predictions, specifically sampled to display a few of the different tropical classes.

\begin{figure*}[t]
    \centering
    \includegraphics[height=0.95\textheight,keepaspectratio]{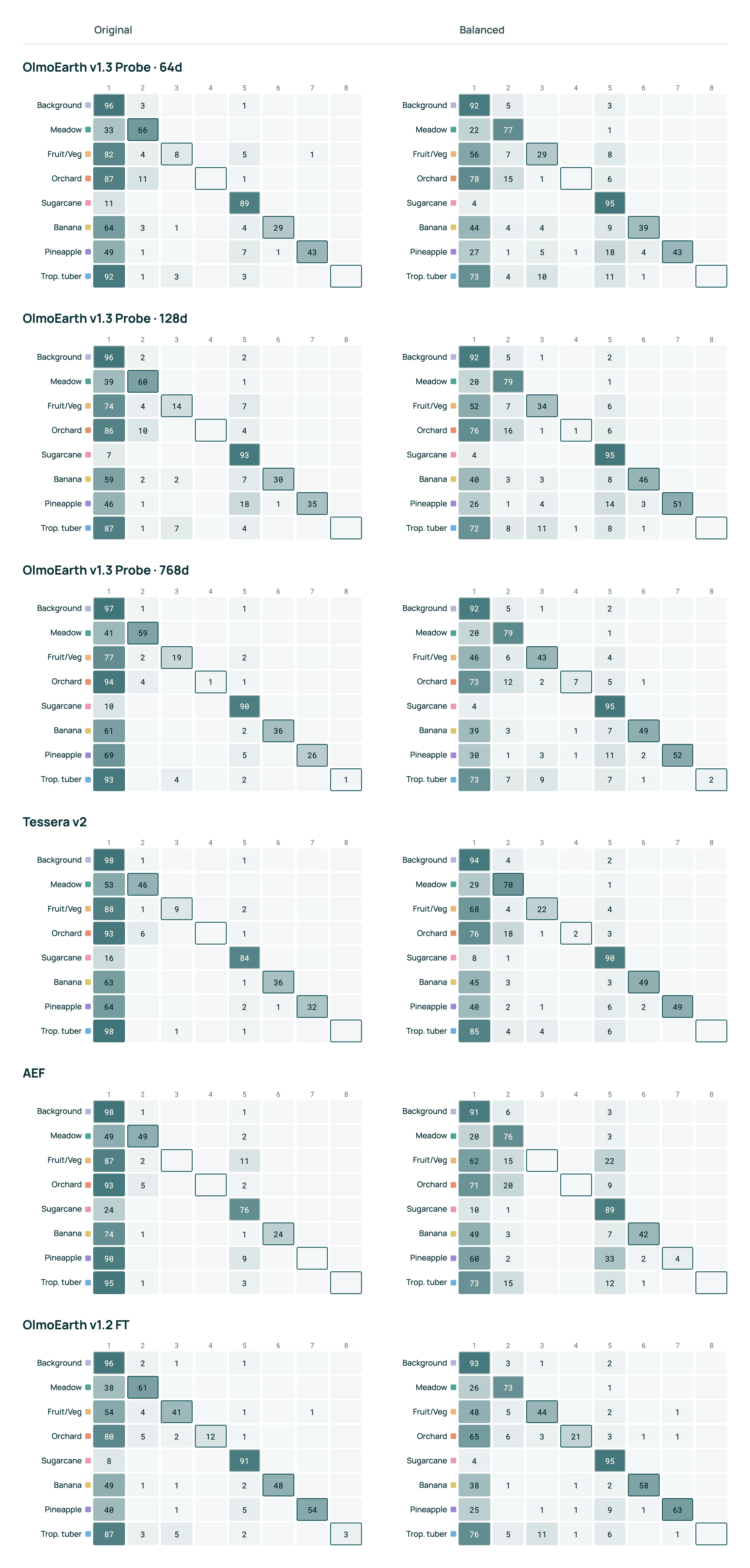}
    \caption{Confusion matrices for methods trained on the imbalanced and balanced train set and evaluated on the 443-window \planteur test split. Rows are ground-truth classes, columns predictions, row-normalized; darker cells carry more mass. Off-diagonal mass sits almost entirely in the Background column.}
    \label{fig:planteur-confusion-matrix-pairs}
\end{figure*}

\paragraph{Confusion Matrices.} Figure~\ref{fig:planteur-confusion-matrix-pairs} shows confusion matrices for \olmoearththree, Tessera v2, AEF, and \olmoearthtwo for the standard setting and the balanced setting, where the training data contained as many Background samples as crop samples.

\clearpage

\end{document}